\documentclass[sigconf]{acmart}

\AtBeginDocument{%
  }

\usepackage{amsmath,amssymb,amsfonts}
\usepackage{graphicx}
\usepackage{textcomp}
\usepackage{xcolor}
\usepackage{booktabs}
\usepackage{multirow}
\usepackage{adjustbox}
\usepackage{mathtools}
\usepackage{float}
\usepackage{makecell}
\usepackage{siunitx}
\usepackage[table]{xcolor}
\usepackage{array}
\usepackage[caption=false,font=footnotesize]{subfig}
\usepackage{bm}

\usepackage{algorithm}
\usepackage{algorithmic}
\usepackage{amsmath}
\usepackage{bm}
\usepackage{booktabs}
\usepackage[normalem]{ulem}
\usepackage{caption}
\usepackage{calc}
\usepackage{enumitem}
\usepackage[above,below]{placeins} 
\usepackage{subcaption}

\copyrightyear{2026}
\acmYear{2026}
\setcopyright{cc}
\setcctype{by}
\acmConference[KDD '26]{Proceedings of the 32nd ACM SIGKDD Conference on Knowledge Discovery and Data Mining V.2}{August 09--13, 2026}{Jeju Island, Republic of Korea}
\acmBooktitle{Proceedings of the 32nd ACM SIGKDD Conference on Knowledge Discovery and Data Mining V.2 (KDD '26), August 09--13, 2026, Jeju Island, Republic of Korea}
\acmDOI{10.1145/3770855.3817960}
\acmISBN{979-8-4007-2259-2/2026/08}

\begin{document}




\title[One Step Closer to Ground Truth: A Multi-Scale Residual-Aware Representation Learning Pipeline]{%
One Step Closer to Ground Truth: A Multi-Scale Residual-Aware
Representation Learning Pipeline for Predicting Time Series Data}


\author{Amrijit Biswas}
\affiliation{%
  \institution{Artificial Intelligence Department, RobotBulls Labs}
  \city{Geneva}
  \country{Switzerland}}
\email{amrijit@robotbulls.com}

\author{Mustafa Kamal}
\affiliation{%
  \institution{Artificial Intelligence Department, RobotBulls Labs}
  \city{Geneva}
  \country{Switzerland}}
\email{mustafa@robotbulls.com}

\author{Robin Krambroeckers}
\affiliation{%
  \institution{Artificial Intelligence Department, RobotBulls Labs}
  \city{Geneva}
  \country{Switzerland}}
\email{robin@robotbulls.com}

\author{M. M. Lutfe Elahi}
\affiliation{%
 \institution{Department of Electrical and Computer Engineering, North South University}
 \city{Dhaka}
 \country{Bangladesh}}
 \email{lutfe.elahi@northsouth.edu}

\author{Sifat Momen}
\affiliation{%
  \institution{Department of Electrical and Computer Engineering, North South University}
  \city{Dhaka}
  \country{Bangladesh}}
  \email{sifat.momen@northsouth.edu}

\author{Nabeel Mohammed}
\affiliation{%
  \institution{Department of Electrical and Computer Engineering, North South University}
  \city{Dhaka}
  \country{Bangladesh}}
\email{nabeel.mohammed@northsouth.edu}

\author{Shafin Rahman}
\affiliation{%
  \institution{Department of Electrical and Computer Engineering, North South University}
  \city{Dhaka}
  \country{Bangladesh}}
\email{shafin.rahman@northsouth.edu}


%
\renewcommand{\shortauthors}{Amrijit Biswas et al.}


\begin{abstract}
Transformer-based models have emerged as leading paradigms in time-series forecasting in recent years, employing self-attention mechanisms to capture long-range dependencies. Despite their success, these single-stage forecasting architectures exhibit persistent systematic residual biases arising from structural discrepancies, unmodeled stochastic components, or inadequate multi-scale temporal representations. This limitation persists when residuals are treated as irreducible noise, precluding adaptive correction of structured error patterns. To address this limitation, we introduce a two-stage, model-agnostic framework that explicitly decouples forecasting and residual learning into distinct stages of representation learning. A base transformer first generates the initial predictions. Subsequently, a dedicated meta-corrector dynamically models structured error patterns across multivariate channels, preserves cross-variable dependencies, and iteratively refines the residual bias of the base transformer. By formalizing this pipeline as a hypothesis space expansion, our framework addresses approximation limitations inherent in single-stage architectures, removes reliance on restrictive assumptions, and enables end-to-end learning of complex error dynamics. Evaluated on eight popular benchmark datasets using established protocols, our approach achieves state-of-the-art performance, with significant improvements in standard metrics (MSE, MAE). The results demonstrate the framework’s ability to mitigate systematic biases and enhance robustness to complex temporal dynamics, advancing the practical applicability of transformer-based forecasting models.
\end{abstract}




\maketitle

\newcommand\kddavailabilityurl{https://doi.org/10.5281/zenodo.20418910}


\section{Introduction}
\label{sec:intro}


\begin{figure}[!t]
    \centering
    \includegraphics[width=\columnwidth]{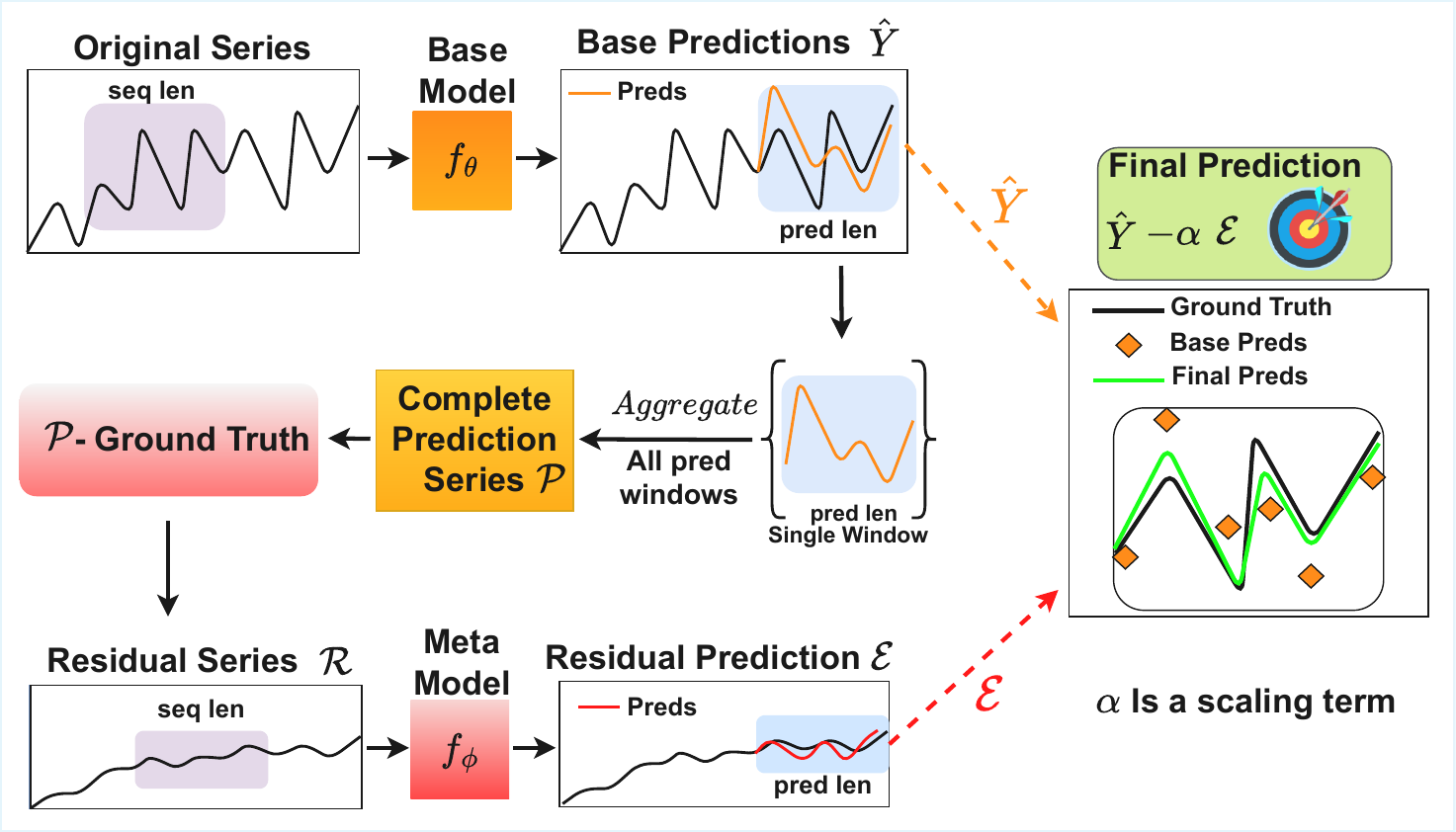}
    \caption{Overview of our proposed pipeline. First, a transformer based model denoted as base model ($f_\theta$) trained on the original series and predictions are generated. From there on, a residual series is calculated by subtracting the ground truth from predictions. After that, another transformer based model named as meta correction model ($f_\phi$) is trained on the residual series to learn the pattern of residual bias made by the base model ($f_\theta$). Finally, the calibrated predictions are obtained by removing the scaled predicted residual bias from base model's predictions. }
    \label{fig:intro}
\end{figure}


Time-series forecasting has garnered substantial interest owing to its wide-ranging applications across diverse domains \cite{morid2023time, hittawe2024time,liu2021pyraformer, zheng2024predicting, aigrain2023gaussian, gonzalez2022prediction, ma2021short}. Transformer-based architectures have markedly advanced long-sequence time-series forecasting (LSTF), with early models introducing specialized attention mechanisms to capture extended dependencies \cite{li2019enhancing, lim2021temporal, zhou2021informer}, while autocorrelation-driven decomposition isolated robust long-term patterns \cite{wu2021autoformer}. To curb complexity, subsequent work reoriented attention along inverted dimensions \cite{liu2023itransformer}, and channel-independent training improved interpretability by modeling each series separately \cite{nie2022time}. More recently, generative AI frameworks enabled end-to-end sequence generation without bespoke forecasting heads \cite{liu2024generative}. Although linear and CNN-based models have shown comparable performance, transformers continue to dominate owing to their superior capacity for modeling long-range dependencies and adaptability across diverse forecasting tasks.

Despite substantial advances in transformer-based LSTF, existing single-stage forecasting frameworks \cite{nie2022time, zeng2023transformers, wen2023transformers, li2024deep, liu2023itransformer, wang2024timexer, liu2024timebridge, qiu2025duet} are inherently constrained in their ability to eliminate systematic bias. (1) Given a fixed model class and training regime, a single-stage predictor frequently reaches an empirical convergence point where further training no longer reduces residuals or improves performance, reflecting an exhaustion of the model’s representational and optimization capacity within that single-stage setup \cite{nakkiran2021deep}. (2) Once converged, the remaining residuals are typically treated as an empirical noise floor, effectively unstructured from the perspective of that model class and training regime \cite{oreshkin2019n}. This treatment precludes any mechanism within the single-stage framework to further refine predictions by modeling residual structure, even when those residuals may contain learnable patterns exploitable by a subsequent modeling stage. (3) Single-stage architectures often exhibit spectral bias \cite{ackaah2023exploring}: state-of-the-art Transformer variants tend to favor certain frequency components while struggling with others, leading to systematic prediction errors. Although linear post‑hoc adjustments \cite{turbe2023evaluation}, boosting ensembles, and ARIMA–LSTM hybrids aim to reduce residual bias, each exhibits fundamental shortcomings in modeling and correcting structured errors over extended horizons.

Building on these observations, we propose a two-stage, residual-aware pipeline that operates on dual scales to overcome single-stage limitations in long-sequence forecasting. Initially, a transformer-based base model generates predictions from the original multivariate series, establishing foundational forecasting capabilities through standard attention mechanisms. Subsequently, residual series are computed as the difference between base predictions and observations, capturing systematic prediction bias that emerges from the base model's inherent approximation limitations. A second transformer model, the meta-corrector, is then trained on these residuals to learn structured error patterns and cross-variable dependencies within the residual space. By operating on different data scales—the base model on original series and meta-corrector on residuals—the framework decouples learning objectives, enabling the base model to focus on raw dynamics while the meta-corrector addresses systematic errors through specialized residual modeling. During inference, the meta-corrector forecasts residual bias, which is subtracted from base predictions to yield corrected long-sequence forecasts with enhanced accuracy and reduced systematic deviation. This two‐stage design enables adaptive bias correction: the meta‐corrector refines the initial predictions by identifying and compensating for systematic deviations arising from complex, multi‐scale trends. Our contributions are as follows:

\begin{itemize}    
    \item We propose a two-stage residual-aware framework that decouples forecasting and residual correction stages at two different scales, enabling existing single-stage transformer architectures to reduce systematic residual biases.  
    \item In our proposed pipeline, model-generated residuals are treated as structured, learnable signals, rather than irreducible noise considered in a single-stage architecture.  
    \item We formalize the residual learning as hypothesis space expansion, where complementary base prediction and meta-corrector stages jointly minimize approximation gaps inherent in single-stage predictors.
    \item We achieved state-of-the-art performance on popular benchmark datasets, with statistically significant improvements in standard forecasting metrics (MSE, MAE) over existing approaches, following established evaluation protocols consistent with prior research. 
\end{itemize}

\section{Related Works}

\subsection{Transformer Based Models} 
Contemporary Transformer‐based forecasters enhance the original architecture by tailoring attention mechanisms to time‐series characteristics. Autoformer \cite{wu2021autoformer} decomposes inputs into trend and seasonal components with auto‐correlation attention, efficiently modeling periodicity but oversimplifying non‐stationary trends. Crossformer \cite{zhang2023crossformer} arranges multivariate data on a 2D time–variable grid and applies temporal then cross‐series attention to bolster inter‐series dependencies, at increased computational cost. PatchTST \cite{nie2022time} reduces quadratic complexity by tokenizing univariate patches and training channel‐independent Transformers, sacrificing cross‐series correlation. The iTransformer \cite{liu2023itransformer} flips inputs to attend over features directly for improved generalization, while TimeXer \cite{wang2024timexer} further integrates exogenous variables via dual attention paths and global tokens, balancing external‐driver accuracy against heightened complexity. TimeBridge \cite{liu2024timebridge} introduces Integrated Attention to suppress short‐term non‐stationarity and Cointegrated Attention to preserve long‐term dependencies, thereby balancing temporal modeling across varying time scales. DUET \cite{qiu2025duet} employs an adaptive RNN framework that matches time‐varying distributions and parameters to capture distinct temporal patterns across variables and periods, offering a unified solution for non‐stationary time series.

\subsection{Linear Models}
Recent linear architectures have likewise demonstrated competitive performance in LSTF tasks. DLinear \cite{zeng2023transformers} challenges prevailing complexity paradigms with a two-linear-layer architecture that decomposes trend and seasonality components, achieving competitive accuracy against advanced transformers while substantially reducing computational overhead; RLinear \cite{li2023revisiting} rigorously analyzes these minimalist components to assess whether the complex mechanisms in state-of-the-art models yield genuine performance gains; TiDE \cite{das2023long} employs a channel-independent MLP with residual blocks and time-derived covariates to deliver adequate long-term forecasts with linear scaling in sequence length, eschewing self-attention, recurrence, and convolution; SparseTSF \cite{lin2024sparsetsf} introduces Cross-Period Sparse Forecasting by downsampling sequences to capture cross-period trends, attaining top-tier results with only 0.92 K parameters for resource-constrained deployment; TimeBase \cite{huangtimebase} emphasizes extreme compactness to provide competitive long-horizon accuracy on edge devices through a minimal-parameter design; and TIMEMIXER++ \cite{wang2024timemixer++} offers a general-purpose framework featuring Multi-Resolution Modeling that adaptively integrates representations across different scales, achieving superior versatility and performance across eight diverse analytical tasks.

\subsection{CNN Based Models}
CNN models have also emerged as prominent approaches for time-series forecasting, owing to their efficient local feature extraction.
SCINet \cite{liu2022scinet} implements a deep multiresolution decomposition architecture with cross-resolution information exchange, yielding superior forecasting performance over Autoformer \cite{wu2021autoformer}, Informer \cite{zhou2021informer}, and base Transformer \cite{vaswani2017attention} by jointly modeling local patterns and global dependencies. TimesNet \cite{wu2022timesnet} transforms 1D series into 2D time–frequency representations via Fourier transforms and parameter-efficient inception blocks with a constant parameter count, delivering strong performance in forecasting, classification, and anomaly detection.
 
However, these methods neglect systematic residual biases by treating them as irreducible noise. To address this shortcoming and further improve forecasting accuracy, we propose our two-stage residual‐learning pipeline.

\section{Methodology}

\begin{figure*}[h]
    \centering
    \includegraphics[width=\textwidth]{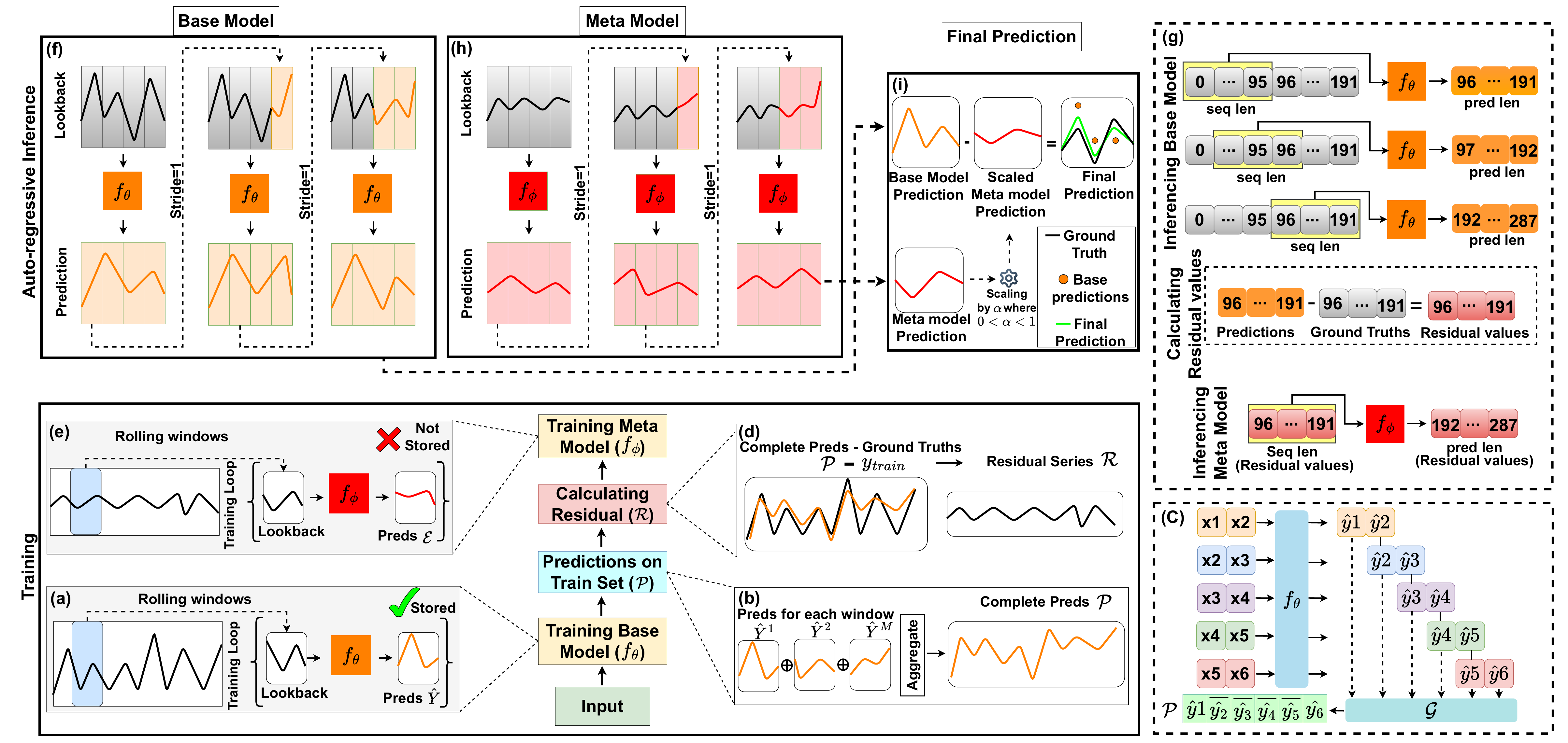}
    \caption{ Overall architecture  of our proposed multi-scale residual-aware representation learning pipeline. (a) First, a base model $f_\theta$ is trained on the original series and predictions are obtained for each given window. (b) The predictions for each window are aggregated to form the complete prediction $\mathcal{P}$. More details in c (c) Predictions are generated in a rolling window manner. Due to the stride value, multiple predictions are generated for a single index. To get a single prediction value per index, the overlapping predictions are aggregated for the same index. (d) The residual series $\mathcal{R}$ is formulated by subtracting the ground truth $y_{train}$ from predictions $\mathcal{P}$.(e) Based on the residual series $\mathcal{R}$, a secondary meta correction model $f_\phi$ is trained to capture the underlying pattern of the residual series $\mathcal{R}$.(f) The base model inference is done in an auto-regressive manner.(g) Given a look-back window of 192 time steps from the original series for forecasting, we first generate predictions using a fixed sequence length of 96 over the available range (indices 96–191). This allows us to compute the residual series $\mathcal{R}$ by subtracting the ground truth from predictions. Using the predictions and ground truth over indices 96–192, we obtain a residual series $\mathcal{R}$ that is aligned with future forecasting targets.(h) Meta model inference also follows the auto-regressive procedure. (i) The final prediction is obtained by integrating the base models’ $f_\theta$ outputs with the meta‑model’s $f_\phi$ predictions after applying an appropriate scaling. } 
    \label{fig:architecture}
\end{figure*}

\subsection{Problem Formulation} 

Let \( X \in \mathbb{R}^{C \times L} \) denote a multivariate time series input with \( C \) channels and \( L \) lookback steps, where \( X = \{\mathbf{x}_1, \mathbf{x}_2, \ldots, \mathbf{x}_L\} \) represents historical observations. Our proposed two-stage residual correction framework first trains a base model \( f_{\theta} \) to minimize forecasting errors. The model is trained on windowed segments \( \{X^{(i)}\}_{i=1}^M \), where \( M \) denotes the number of windows, to predict the next \( H \) future time steps for each window. Formally, we parameterize the base model \( f_{\theta} \) to learn the mapping:
\begin{equation}
\widehat{\mathbf{\mathbf{Y}}}_{L+1:L+H} = f_{\theta}(\mathbf{X}_{1:L}).
\label{eq:prediction}
\end{equation}
where each \( \mathbf{X} \in \mathbb{R}^{C \times L} \) represents a windowed segment of the input series and \( \widehat{\mathbf{Y}} \in \mathbb{R}^{C \times H} \) denotes the corresponding predicted horizon. These predictions are aggregated into a consolidated series \( \mathcal{P} = \mathcal{G}(\{\hat{\mathbf{Y}}^{(i)}\}_{i=1}^{M}) \). The residual series is then computed as \( \mathcal{R} = \mathcal{P} - \mathcal{Y}^{\text{train}} \), where \( \mathcal{Y}^{\text{train}} \) represents the complete set of actual ground truths corresponding to the training period. Subsequently, a secondary meta model, \( f_{\phi} \) learns to predict these residuals following the same windowed training procedure, operating on residual windows \( \{\mathcal{R}^{(i)}\}_{i=1}^M \). The residual meta correction model \( f_{\phi} \) processes windowed residual segments \( \mathcal{R}_{1:L} \) to predict future residuals:
\begin{equation}
{\mathbf{\mathcal{E}}}_{L+1:L+H} = f_{\phi}(\mathcal{R}_{1:L})
\end{equation}
During testing, we adopt an autoregressive strategy for both models. The base model \( f_{\theta} \) predicts from a fixed-length input window \( X^i \in \mathbb{R}^{C \times L} \), producing the base forecast \( \hat{Y}^i = f_{\theta}(X^i) \). The residual correction model \( f_{\phi} \) is first applied to the initial window \( R^i \in \mathbb{R}^{C \times L} \) aligned with \( X^i\) to estimate the residual \( \mathcal{E}^1 \in \mathbb{R}^{C \times H} \). 

For each subsequent window \( i > 1 \),  the residual input \( R^i \in \mathbb{R}^{C \times L} \) is constructed autoregressively from previous predictions \( \{\mathcal{E}^{j}\}_{j=1}^{i-1} \), and used to compute \( \mathcal{E}^i = f_{\phi}(R^i) \). The final forecast for the \( i \)-th window is given by:
\begin{equation}
    \hat{Y}^{i}_{\text{final}} = f_{\theta}(X^i) - \alpha \cdot f_{\phi}(\mathcal{R}^i),
\end{equation}
where \( \alpha \in (0,1) \) is a scaling factor that prevents over correction.

\noindent\textbf{Challenges:}

We formalize the existing challenges of a single-stage forecasting model below: 

\noindent\textit{\textbf{(1) Persistent Systematic Residual Bias:}} Despite achieving convergence, single-stage models may still exhibit non-trivial structured errors in their forecasts. Real-world time series often contain overlapping patterns at different scales (trends, seasonal cycles, sub-daily variations, etc.). Single-stage models tend to learn the dominant trend but miss subtler multi-scale structures. In practice, any unmodeled component of the signal remains as correlated “residual” error. After the dominant temporal patterns have been learned by the model, the leftover systematic discrepancies (non-white residuals) persist unchecked. Standard single-stage architectures have no built-in mechanism to disentangle or correct these patterned residuals, so they propagate as persistent bias in the outputs.

\noindent\textit{\textbf{(2) Spectral Bias:}} Deep forecasting models may prioritize low-frequency (smooth) components of the time series, potentially overlooking high-frequency details. Existing time-series Transformers “favor low frequency and have difficulty capturing high-frequency information,” a phenomenon called as spectral bias \cite{ackaah2023exploring}. High‑frequency fluctuations in time series often correspond to critical phenomena such as abrupt regime shifts, transient anomalies, or short‑term volatility. However, models inherently prioritize fitting smooth, low frequency trends both because these components carry the bulk of the variance in a typical loss function and because their architectures (e.g. self‑attention mechanisms) exhibit a built‑in low‑pass bias—leaving high-frequency components systematically underrepresented. This consequently gives rise to unmodeled residual bias.

\noindent\textit{\textbf{(3) Epistemic Uncertainty Propagation:}}
In single‐stage forecasting, any mismatch between the true data‐generating process and the model’s fixed hypothesis class produces an irreducible approximation error, epistemic uncertainty, that the framework simply treats as random noise. Because there is no mechanism to identify or correct this structured error, it persists and compounds in downstream predictions, undermining robustness when the underlying dynamics lie outside the model’s representational capacity.

\subsection{Multi-Scale Residual-Aware Representation Learning Pipeline} 

We propose a multi-scale residual-aware forecasting framework that decomposes the prediction task through two specialized models operating on: (1) a base predictor $f_\theta$ trained on raw time series observations $X^{train}$, and (2) a meta-corrector $f_\phi$ that learns to model structured residuals $\mathcal{R} = \mathcal{P} - X^{train}$. The architecture is motivated by the key hypothesis that conventional single-stage residuals contain systematic, learnable residual patterns that can be captured through explicit modeling. By combining both components through $\hat{Y}^{i}_{\text{final}} = f_{\theta}(X^i) - \alpha \cdot f_{\phi}(R^i)$, our framework simultaneously reduces epistemic uncertainty while preserving the interpretability of the base model's predictions. This dual-scale approach enables hierarchical feature learning, wherein the base model \( f_\theta \) captures dominant temporal patterns, while the residual model \( f_\phi \) targets structured components within the residuals that are otherwise treated as random noise by \( f_\theta \). The subsequent sections elaborate on the training procedure, and evaluation metrics. 


\subsubsection{\textbf{Training the Base Model}}
The base model \( f_{\theta} \) is trained on the multivariate input $X^{train}$, following iTransformer ~\cite{liu2023itransformer}. Given an input time series \( X = \{\mathbf{x}_1, \mathbf{x}_2, \ldots, \mathbf{x}_L\} \in \mathbb{R}^{C \times L} \), the model processes it in windows \( \{ X^{(i)}\}_{i=1}^M \), where each \( X^{(i)} \) consists of \( L \) historical time steps across \( C \) channels. The encoder employs multi-variate self-attention \cite{liu2023itransformer} to capture temporal dependencies, followed by position-wise feedforward layers. Layer normalization \cite{vaswani2017attention} and residual connections are applied for stable training. The output of the encoder is projected into the future horizon \( H \) via a linear layer, producing predictions as shown in Equation ~\eqref{eq:prediction}
\noindent
During training, we minimize the mean squared error (MSE) between the model's predictions \( \widehat{\mathbf{Y}}^i \) and the corresponding ground truth values \( {\mathbf{Y}}^i \) for each window. These window-level predictions are then aggregated to form a continuous output series, denoted as \( \mathcal{P} \).

\subsubsection{\textbf{Prediction Aggregation and Residual Series Construction}}
Given a sliding lookback window of size \( L \), the model generates a sequence of \( H \) future values for each input window. Consecutive windows produce predictions that partially overlap. Specifically, predictions for time steps \( L+2 \) through \( L+H \) in one window coincide with predictions for earlier steps in the subsequent windows. As a result, each future time index beyond \( L+1 \) receives multiple predicted values from different windows. This redundancy necessitates an aggregation function $\mathcal{G}$ to construct a unified prediction sequence \( \mathcal{P} \), particularly to ensure consistency when computing the residual sequence \( \mathcal{R} \). 

The aggregation function $\mathcal{G}$ combines overlapping predictions by averaging all predictions associated with a given time index \( i \), yielding the final aggregated forecast:
\begin{equation}
    \mathcal{P} = \mathcal{G}(\{\hat{\mathbf{Y}}^{(i)}\}_{i=1}^{M}), \quad \text{where} \quad P_t = \frac{1}{|\mathcal{I}_t|} \sum_{i \in \mathcal{I}_t} \hat{y}^{(i)}_t
    \label{eq:aggregation}
\end{equation}
Here, $\hat{\mathbf{Y}}^{(i)} \in \mathbb{R}^{C \times L}$ is the output of the $i$-th window, $\mathcal{I}_t$ denotes the set of windows covering overlapping index $t$, and $\mathcal{P}$ is the final sequence obtained by averaging overlapping predictions at each index. This process is similar to the aggregation mechanism used in ensemble learning methods such as Random Forests~\cite{zhou2023comparative}, as illustrated in Fig. ~\ref{fig:architecture}(C).

The aggregated prediction series \(\mathcal{P}\), obtained by aggregating overlapping window predictions, does not contain values corresponding to the initial \(L\) time steps of the training set. Therefore, during residual calculation, we compute

\begin{equation}
    \mathcal{R} = \mathcal{P} - X^{\text{train}}_{L+1:T}
    \label{eq:residual}
\end{equation}
where the segment \(X^{\text{train}}_{L+1:T}\) excludes the first \(L\) values to align with \(\mathcal{P}\), and \(T\) denotes the total length of the training set.

\subsubsection{\textbf{Training Secondary Meta Corrector Model}}
The meta-corrector model \( f_{\phi} \) is trained on the residual series \(\mathcal{R}\), which captures the structured errors generated from the base model’s predictions. Given a residual window \(\mathcal{R}_{1:L} \in \mathbb{R}^{C \times L}\), the model predicts the future residual values over the horizon \(H\) as:
\begin{equation}
    {\mathbf{\mathcal{E}}}_{L+1:L+H} = f_{\phi}(\mathcal{R}_{1:L}),
    \label{eq:residual_preds}
\end{equation}
where \({\mathbf{\mathcal{E}}}_{L+1:L+H} \in \mathbb{R}^{C \times H}\) denotes the predicted residuals. The meta-corrector is rigorously optimized using the Huber loss \cite{gokcesu2021generalized}, which effectively balances robustness against continual sign and magnitude altering residual patterns, enabling more stable and accurate correction of the base model's errors.

\subsubsection{\textbf{Final Prediction}}

To generate test-set predictions in our two-stage forecasting framework, we employ an autoregressive strategy. The \textit{base model}, denoted by \( f_{\theta} \), operates on a fixed-length input window of size \( L \). However, to produce the initial residual prediction, a longer input window of length \( 2L \) is required to provide additional context for error correction. This extended input, \( \tilde{X}^1 \in \mathbb{R}^{C \times 2L} \), contains the same \( L \) time steps used by the base model as its second half, ensuring alignment between the inputs of both models.

At inference time, due to the absence of ground truth labels, the model constructs the initial residual input from the extended input sequence. Specifically, the first half of \( \tilde{X}^1 \), denoted \( \tilde{X}^1_{1:L} \), is passed to the base model \( f_{\theta} \) to obtain a prediction, and the second half, \( \tilde{X}^1_{L+1:2L} \), is treated as target. The difference between these two components defines the initial residual input to the residual model. We compute the first residual series input by $\mathcal{R}^1 = \tilde{X}^1_{L+1:2L} - \hat{Y}_{L+1:2L} $. This residual input is then passed to \( f_{\phi} \) to generate the residual prediction \( \mathcal{E}^1 \). Subsequent residual inputs are constructed autoregressively, each new input is formed by shifting forward one stride and feeding in the previously predicted residuals \( \{\mathcal{E}^j\}_{j=1}^{i-1} \) to compute \( \mathcal{E}^i = f_{\phi}(R^i) \).

However, subtracting the meta-model’s output from the base model’s prediction may sometimes result in over-correction. In a two-stage forecasting framework, the meta-corrector $f_\phi$ must accurately predict both the sign and magnitude of the base model's residuals. If the residual estimate ${\mathcal{E}}$ has the wrong sign or an incorrect magnitude, its calibration can degrade rather than improve the forecast. To address this forecast stability and mitigate the impact of erroneous residuals, we incorporate a scaling factor \( \alpha \in (0,1) \) to scale the meta-corrector’s residual predictions. Inspired by prior work \cite{deghfel2024new, friedman2001greedy, chen2016xgboost, huang2024improved, li2025rethinking, dey2024sparse}, this scaling adjusts the final prediction as follows:
\begin{equation}
    \hat{Y}^{i}_{\text{final}} = \hat{Y}^i - \alpha \cdot \mathcal{E}^i = f_{\theta}(X^i) - \alpha \cdot f_{\phi}(R^i)
    \label{eq:final}
\end{equation}


\begin{algorithm}[!t]
\caption{Proposed Multi-Scaled Residual-Aware pipeline}
\begin{algorithmic}[1]
\REQUIRE Input series $\mathbf{X}\in\mathbb{R}^{C\times L}$; Base model \( f_{\theta} \); residual correction model \( f_{\phi} \); Lookback steps $L$; Future prediction length $H$; Number of channels $C$; Number of windows $M$

\STATE Train the \( f_{\theta} \) using windowed segments \( \{X^{(i)}\}_{i=1}^M \) inputs

\STATE $\{\hat{\mathbf{Y}}^{(i)}\}_{i=1}^{M} = f_{\theta}({\{X^{(i)}\}_{i=1}^M})$

\STATE Concatenate window predictions into a consolidated sequence.

\STATE \( \mathcal{P} = \mathcal{G}(\{\hat{\mathbf{Y}}^{(i)}\}_{i=1}^{M}) \)

\STATE Generate the residual series by subtracting $\mathcal{Y}^{\text{train}}$ from $\mathcal{P}$.

\STATE \( \mathcal{R} = \mathcal{P} - \mathcal{Y}^{\text{train}} \)

\STATE Train residual correction model \( f_{\phi} \) using residuals \( \{\mathcal{R}^{(i)}\}_{i=1}^M \).
\STATE  \({\mathbf{\mathcal{E}^i}}= f_{\phi}(\{\mathcal{R}^{(i)}\}_{i=1}^M)\)

\STATE Both models' outputs are combined and scaled to produce the final forecast.

\STATE $\hat{Y}^{i}_{\text{final}} = f_{\theta}(X^i) - \alpha \cdot f_{\phi}(\mathcal{R}^i)$

\RETURN $\hat{Y}^{i}_{\text{final}}$

\end{algorithmic}
\end{algorithm}

\subsection{Implementation confrontations}

\noindent\textbf{Addressing Overfitting and Data Leakage} 

To ensure methodological rigor and guard against overfitting and data leakage, we adopt a two‑stage training scheme with strictly isolated datasets and no shared parameters. In the first stage, the base model 
$f_\theta$ is trained on the original dataset to learn the underlying feature dynamics. Upon completion, its weights are frozen and its predictions are used to compute the residual dataset. In the second stage, the meta‑correction model $f_\phi$ is trained exclusively on this residual dataset, with $f_\theta$ remaining inactive and without access to any of its weights or statistical properties, thereby eliminating any risk of information leakage.

\noindent\textbf{Recurring alternating positive and negative trends observed in the residual series.} In the fluctuation dataset, feature signs change frequently, obscuring underlying patterns and increasing the risk of fitting noise. Accurately predicting both sign and magnitude is therefore critical, as errors propagate into the final output and can degrade performance below that of the original model. To address this, we substitute the original Mean Squared Error loss in iTransformer \cite{liu2023itransformer} with the Huber loss, which more robustly penalizes both large and small deviations. Empirically, this modification yields improved accuracy in fluctuation prediction relative to MSE.
\begin{equation}
    Huber Loss_{\delta}(a) =
\begin{cases} 
\frac{1}{2} a^2, & \text{if } |a| \leq \delta \\
\delta (|a| - \frac{1}{2} \delta), & \text{if } |a| > \delta
\end{cases}
\end{equation}
Where, $\alpha= \hat y- y$ represents the difference between the predicted and true values and $\delta$ is a hyperparameter that controls the threshold for the error term, determining the transition point between the quadratic and linear regions of the loss function.

In instances where the predicted and true fluctuations have opposite signs (e.g., $+x$ versus $-x$) the absolute residual $|a|$ exceeds the Huber threshold $\delta$, thereby invoking its linear regime: $L(a) = \delta\!\bigl(\lvert a\rvert - \tfrac{1}{2}\,\delta\bigr)$. This piecewise linear penalty furnishes a robust yet stable gradient signal for sign‐error correction, avoiding the unbounded quadratic growth characteristic of MSE, and imposes a clear separation between sign inversions and minor deviations, ultimately enhancing generalization.

\noindent\textbf{Preserving magnitude divergence during model training on both the original and residual series.} We propose a two‑stage forecasting framework that explicitly preserves the scale disparity between the high‑magnitude original series and the low‑magnitude residual series. Although both series undergo a uniform standard scaling prior to model input, this procedure compresses their natural magnitude difference and risks causing residual corrections to overshoot instead of refining the base forecast. To mitigate this effect, we inverse‑transform the model outputs for both the original and residual series back to their respective native scales before applying the residual correction (i.e., subtracting the residual forecast from the base forecast). Finally, we reapply the standard scaling to the combined prediction to facilitate direct performance comparison with existing benchmarks.

\section{Experiments}

\subsection{Setup}
\label{sub:setup}

\noindent\textbf{Dataset:}

To rigorously assess the generalizability of our proposed pipeline in diverse application areas, we conducted experiments on eight LSTF benchmark datasets that are the most widely used: the four subsets of the ETT dataset \cite{zhou2021informer} for electricity trading loads; the Electricity (ECL) dataset \cite{li2019enhancing} for household power consumption; the Traffic dataset \cite{wu2022timesnet} for urban vehicular flow; the Weather dataset \cite{zhou2021informer} for meteorological variables; and the Exchange Rate dataset \cite{wu2021autoformer} for interbank currency series.

\noindent\textbf{Baselines:}
We include eleven latest state‑of‑the‑art long‑sequence time series forecasting models: TimeMixer++ \cite{wang2024timemixer++}, SparseTSF (ICML 2025) \cite{lin2024sparsetsf}, TimeBase (ICML 2025) \cite{huangtimebase}, DUET (KDD 2025) \cite{qiu2025duet}, TimeBridge (ICML 2025) \cite{liu2024timebridge}, TimeXer (NeurIPS 2024) \cite{wang2024timexer}, iTransformer (ICLR 2024) \cite{liu2023itransformer}, GPHT (KDD 2024) \cite{liu2024generative}, Amplifier (AAAI 2025) \cite{fei2025amplifier} and xPatch (AAAI 2025) \cite{stitsyuk2025xpatch}.

\noindent\textbf{Implementation Details}
In our multivariate time series forecasting experiment, we used iTransformer \cite{liu2023itransformer} as our foundational transformer-based model. We maintain the same fixed input sequence length of 96 and the four prediction lengths \{96, 192, 336, 720\} as in iTransformer. During training and inference, data shuffling is disabled, and a rolling window with stride 1—identical to iTransformer—is applied for both models so that each batch is drawn sequentially from the beginning to the end of the dataset, thereby preserving temporal alignment between input sequences and their corresponding predictions. The batch size is dataset-dependent, with a default value of 32. Following TFB’s “Drop Last” trick \cite{qiu2024tfb}, the final (possibly smaller) batch is retained in both training and inference. The base model is trained using the mean squared error (MSE) loss function, whereas the meta-correction model employs the Huber loss. Both models are trained for 10 epochs using the Adam optimizer, with early stopping (patience = 3). All experiments were conducted on a single NVIDIA RTX 4090 GPU. Final performance is evaluated in a window-wise manner, consistent with the testing protocols adopted in prominent prior works.

\subsection{Results}

\subsubsection{\textbf{Evaluating Residual Prediction Performance}}

Accurate residual prediction is critically important in our pipeline, where modest improvements yield substantial overall performance gains. Conventional time-series benchmarks typically employ scaled inputs that mask prediction error magnitudes and produce artificially deflated MSE and MAE values. To ensure meaningful evaluation, we report all performance metrics on the raw data scale in the Appendix Table~\ref{tab:full_fluctuation_performance}, revealing substantially different error magnitudes that offer a more realistic assessment of predictive performance. Our empirical evaluation demonstrates that Huber loss function consistently outperforms MSE loss for residual forecasting. While their results occasionally coincide, MSE never exceeds Huber in performance. On the ETT datasets, Huber's advantage is modest yet consistent. On datasets with high-magnitude target values (e.g., Electricity and Weather), the gap in raw-scale MSE and MAE is substantial and would be entirely masked by data scaling. Even on the lower-magnitude Traffic dataset, where residual errors exhibit high sensitivity, the improvements achieved under Huber loss prevent error propagation and preserve downstream performance.


\begin{table*}[h]
  \centering
  \caption{Complete results of the long-term forecasting task, evaluating various models across prediction lengths $S \in \{96,192,336,720\}$. Input length is fixed at 96.}
  \scriptsize
  \setlength{\tabcolsep}{2pt}  
  \renewcommand{\arraystretch}{1.1}
  \resizebox{\textwidth}{!}{%
    \begin{tabular}{@{}c|c*{12}{|cc}@{}}
      \toprule
      \multirow{2}{*}{\rotatebox{90}{\textbf{Dataset}}}
        & \multirow{2}{*}{\makecell[l]{\textbf{Models}\\\textbf{Metric}}}
        & \multicolumn{2}{c}{\makebox[2pt][c]{\textbf{Ours}}}
        & \multicolumn{2}{c}{\makebox[2pt][c]{\shortstack{\textbf{TimeXer}\\\cite{wang2024timexer}}}}
        & \multicolumn{2}{c}{\makebox[2pt][c]{\shortstack{\textbf{iTransformer}\\\cite{liu2023itransformer}}}}
        & \multicolumn{2}{c}{\makebox[2pt][c]{\shortstack{\textbf{GPHT}\\\cite{liu2024generative}}}}
        & \multicolumn{2}{c}{\makebox[2pt][c]{\shortstack{\textbf{TimeMixer$++$}\\\cite{wang2024timemixer++}}}}
        & \multicolumn{2}{c}{\makebox[2pt][c]{\shortstack{\textbf{SparseTSF}\\\cite{lin2024sparsetsf}}}}
        & \multicolumn{2}{c}{\makebox[2pt][c]{\shortstack{\textbf{TimeBase}\\\cite{huangtimebase}}}}
        & \multicolumn{2}{c}{\makebox[2pt][c]{\shortstack{\textbf{DUET}\\\cite{qiu2025duet}}}}
        & \multicolumn{2}{c}{\makebox[2pt][c]{\shortstack{\textbf{TimeBridge}\\\cite{liu2024timebridge}}}}
        & \multicolumn{2}{c}{\makebox[2pt][c]{\shortstack{\textbf{TimeFilter}\\\cite{hu2025timefilter}}}}
        & \multicolumn{2}{c}{\makebox[2pt][c]{\shortstack{\textbf{Amplifier}\\\cite{fei2025amplifier}}}}
        & \multicolumn{2}{c}{\makebox[2pt][c]{\shortstack{\textbf{xPatch}\\\cite{stitsyuk2025xpatch}}}} \\
      \cmidrule(lr){3-4}  \cmidrule(lr){5-6}  \cmidrule(lr){7-8}  \cmidrule(lr){9-10}
      \cmidrule(lr){11-12} \cmidrule(lr){13-14} \cmidrule(lr){15-16} \cmidrule(lr){17-18}
      \cmidrule(lr){19-20} \cmidrule(lr){21-22} \cmidrule(lr){23-24} \cmidrule(lr){25-26}
      & & MSE & MAE & MSE & MAE & MSE & MAE & MSE & MAE & MSE & MAE & MSE & MAE & MSE & MAE & MSE & MAE & MSE & MAE & MSE & MAE & MSE & MAE & MSE & MAE \\
      \midrule


          \multirow{5}{*}{\textbf{\rotatebox{90}{ETTh1}}} 
        & 96  & 0.367 & \textbf{\textcolor{red}{0.311}} & 0.382 & 0.403 & 0.386 & 0.405 & 0.363 & \textbf{\textcolor{blue}{0.382}} & 0.361 & 0.403 & 0.362 & 0.389 & 0.349 & 0.384 & \textbf{\textcolor{blue}{0.352}} & 0.384 & \textbf{\textcolor{red}{0.350}} &  0.389 & 0.370 & 0.394 & 0.371 & 0.392 & 0.376 & 0.386 \\

        & 192 & 0.428 & \textbf{\textcolor{red}{0.337}} & 0.429 & 0.435 & 0.441 & 0.436 & 0.405 & \textbf{\textcolor{blue}{0.408}} & 0.416 & 0.441 & 0.404 & 0.412 & \textbf{\textcolor{red}{0.387}} & 0.410 & 0.398 & 0.409 & \textbf{\textcolor{blue}{0.388}} & 0.414 & 0.413 & 0.420 & 0.426 & 0.422 & 0.417 & 0.407 \\
    
        & 336 & 0.475 & \textbf{\textcolor{red}{0.355}} & 0.468 & 0.448 & 0.487 & 0.458 & 0.430 & \textbf{\textcolor{blue}{0.423}} & 0.430 & 0.434 & 0.435 & 0.428 & \textbf{\textcolor{red}{0.408}} & 0.418 & \textbf{\textcolor{blue}{0.414}} & 0.426 & \textbf{\textcolor{red}{0.408}} & 0.430 & 0.450 & 0.440 & 0.448 & 0.434 & 0.449 & 0.425 \\
     
        & 720 & 0.464 & \textbf{\textcolor{red}{0.373}} & 0.469 & 0.461 & 0.503 & 0.491 & \textbf{\textcolor{red}{0.414}} & \textbf{\textcolor{blue}{0.435}} & 0.467 & 0.451 & \textbf{\textcolor{blue}{0.426}} & 0.448 & 0.439 & 0.446 & 0.429 & 0.455 & 0.443 & 0.463 & 0.448 & 0.457 & 0.476 & 0.464 & 0.470 & 0.456 \\

        \midrule

        \multirow{5}{*}{\textbf{\rotatebox{90}{ETTh2}}} 
        & 96  & \textbf{\textcolor{red}{0.054}} & \textbf{\textcolor{red}{0.144}} & 0.286 & 0.338 & 0.297 & 0.349 & 0.296 & 0.340 & 0.276 & 0.328 & 0.294 & 0.346 & 0.292 & 0.345 & 0.270 & 0.336 & 0.271 & 0.331 & 0.283 & 0.337 & 0.279 & 0.337 & \textbf{\textcolor{blue}{0.233}} & \textbf{\textcolor{blue}{0.300}} \\
    
        & 192 & \textbf{\textcolor{red}{0.066}} & \textbf{\textcolor{red}{0.161}} & 0.363 & 0.389 & 0.380 & 0.400 & 0.363 & 0.384 & 0.342 & 0.379 & 0.340 & 0.377 & 0.339 & 0.387 & 0.332 & 0.374 &  0.335 & 0.370 & 0.362 &  0.392 & 0.359 & 0.389 & \textbf{\textcolor{blue}{0.291}} & \textbf{\textcolor{blue}{0.338}} \\
        
        & 336 & \textbf{\textcolor{red}{0.078}} & \textbf{\textcolor{red}{0.178}} & 0.414 & 0.423 & 0.428 & 0.432 & 0.392 & 0.410 & 0.346 & 0.398 & 0.360 & 0.398 & 0.358 & 0.410 & 0.353 & 0.397 & 0.371 & 0.402 & 0.404 & 0.424 & 0.377 & 0.406 & \textbf{\textcolor{blue}{0.344}} & \textbf{\textcolor{blue}{0.377}} \\
        
        & 720 & \textbf{\textcolor{red}{0.078}} & \textbf{\textcolor{red}{0.183}} & 0.408 & 0.432 & 0.427 & 0.445 & 0.407 & 0.427 & 0.392 & \textbf{\textcolor{blue}{0.415}} & 0.383 & 0.425 & 0.400 & 0.448 & \textbf{\textcolor{blue}{0.382}} & 0.425 & 0.387 & 0.425 & 0.407 &  0.433 & 0.420 & 0.432 & 0.407 & 0.427 \\
        
        \midrule

        \multirow{5}{*}{\textbf{\rotatebox{90}{ETTm1}}}
        
        & 96  & 0.293 & \textbf{\textcolor{red}{0.273}} & 0.318 & 0.356 & 0.334 & 0.368 & 0.291 & 0.339 & 0.310 &  0.334 & 0.314 & 0.359 &  0.311 & 0.351 & \textbf{\textcolor{red}{0.279}} & \textbf{\textcolor{blue}{0.333}} & \textbf{\textcolor{blue}{0.284}} & 0.337 &  0.313 & 0.354 & 0.316 & 0.355 & 0.311 & 0.346 \\
   
        & 192 & 0.343 & \textbf{\textcolor{red}{0.296}} & 0.362 & 0.383 & 0.377 & 0.391  & 0.337 & 0.368 & 0.348 & 0.362 & 0.348 & 0.376 & 0.338 & 0.371 & \textbf{\textcolor{blue}{0.320}} & \textbf{\textcolor{blue}{0.358}} & \textbf{\textcolor{red}{0.317}} & 0.367 &  0.356 & 0.380 & 0.361 & 0.381 & 0.348 & 0.368 \\

        & 336 & 0.383 & \textbf{\textcolor{red}{0.318}} & 0.395 & 0.407 & 0.426 & 0.420 & 0.377 & 0.393 & 0.376 & 0.391 & 0.368 & 0.386 & 0.364 & 0.386 & \textbf{\textcolor{red}{0.348}} & \textbf{\textcolor{blue}{0.377}} & \textbf{\textcolor{blue}{0.361}} & 0.394 &  0.386 &  0.402 & 0.393 & 0.402 & 0.388 & 0.391 \\

        & 720 & 0.465 & \textbf{\textcolor{red}{0.354}} & 0.452 & 0.441 & 0.491 & 0.459 & 0.452 & 0.433 & 0.440 & 0.423 & 0.419 & 0.413 & \textbf{\textcolor{blue}{0.413}} & 0.414 & \textbf{\textcolor{red}{0.405}} & \textbf{\textcolor{blue}{0.408}} & \textbf{\textcolor{blue}{0.413}} & 0.418 & 0.452 & 0.437 & 0.455 & 0.440 & 0.461 & 0.430 \\

        \midrule

        \multirow{5}{*}{\textbf{\rotatebox{90}{ETTm2}}} 
        & 96  & \textbf{\textcolor{red}{0.037}} & \textbf{\textcolor{red}{0.116}} & 0.171 & 0.256 & 0.180 & 0.264 & 0.170 & 0.250 & 0.170 & 0.245 & 0.167 & 0.259 & 0.162 & 0.256 & 0.161 & 0.248 & \textbf{\textcolor{blue}{0.157}} & \textbf{\textcolor{blue}{0.243}} &  0.169 & 0.255 & 0.176 & 0.258 & 0.164 & 0.248 \\
    
        & 192 & \textbf{\textcolor{red}{0.048}} & \textbf{\textcolor{red}{0.130}} & 0.237 & 0.299 & 0.250 & 0.309 & 0.230 & 0.291 & 0.229 & 0.291 & 0.219 & 0.297 & 0.218 & 0.293 & \textbf{\textcolor{blue}{0.214}} & 0.286 & 0.217 & \textbf{\textcolor{blue}{0.285}} & 0.235 & 0.299 & 0.239 & 0.300 & 0.230 & 0.291  \\

        & 336 & \textbf{\textcolor{red}{0.059}} & \textbf{\textcolor{red}{0.146}} & 0.296 & 0.338 & 0.311 & 0.348 & 0.285 & 0.327 & 0.303 & 0.343 & 0.271 & 0.330 & 0.270 & 0.328 & \textbf{\textcolor{blue}{0.267}} & \textbf{\textcolor{blue}{0.321}} & 0.269 & \textbf{\textcolor{blue}{0.321}} & 0.293 & 0.336 & 0.297 & 0.338 & 0.292 & 0.331 \\
    
        & 720 & \textbf{\textcolor{red}{0.074}} & \textbf{\textcolor{red}{0.166}} & 0.392 & 0.394 & 0.412 & 0.407 & 0.380 & 0.386 & 0.373 & 0.399 & 0.353 & 0.380 & 0.352 & 0.380 & \textbf{\textcolor{blue}{0.348}} & \textbf{\textcolor{blue}{0.374}} & \textbf{\textcolor{blue}{0.348}} & 0.378 & 0.390 &  0.393 & 0.393 & 0.396 & 0.381 & 0.383  \\

        \midrule

        \multirow{5}{*}{\textbf{\rotatebox{90}{Electricity}}} 
        & 96  & \textbf{\textcolor{red}{0.017}} & \textbf{\textcolor{red}{0.013}} & 0.140 & 0.242 & 0.148 & 0.240 & 0.128 & 0.219 & 0.135 & 0.222 & 0.139 & 0.239 &  0.139 & 0.231 &  0.128 & 0.219 & \textbf{\textcolor{blue}{0.120}} & \textbf{\textcolor{blue}{0.214}} & 0.133 & 0.230 &  0.147 & 0.243 & 0.159 & 0.244  \\
  
        & 192 & \textbf{\textcolor{red}{0.023}} & \textbf{\textcolor{red}{0.014}} & 0.157 & 0.256 & 0.162 & 0.253& 0.146 & 0.236 & 0.147 & \textbf{\textcolor{blue}{0.235}} & 0.155 & 0.250 & 0.153 & 0.245 & 0.145 & \textbf{\textcolor{blue}{0.235}} & \textbf{\textcolor{blue}{0.142}} & 0.237 & 0.154 & 0.248 & 0.157 & 0.251 & 0.160 & 0.248  \\
    
        & 336 & \textbf{\textcolor{red}{0.038}} & \textbf{\textcolor{red}{0.019}} & 0.176 & 0.275 & 0.178 & 0.269 & 0.165 & 0.255 & 0.164 & \textbf{\textcolor{blue}{0.245}} & 0.171 & 0.265 & 0.169 & 0.262 & 0.163 & 0.255 & \textbf{\textcolor{blue}{0.156}} & 0.252 &  0.162 & 0.261 & 0.174 & 0.269 & 0.182 & 0.267  \\

        & 720 & \textbf{\textcolor{red}{0.045}} & \textbf{\textcolor{red}{0.026}} & 0.211 & 0.306 & 0.225 & 0.317 & 0.207 & 0.292 & 0.212 & 0.310 & 0.208 & 0.300 & 0.207 & 0.294 & 0.193 & 0.281 &  \textbf{\textcolor{blue}{0.179}} & \textbf{\textcolor{blue}{0.278}} & 0.184 & 0.284 & 0.206 & 0.296 & 0.216 & 0.298 \\

        \midrule

        \multirow{5}{*}{\textbf{\rotatebox{90}{Traffic}}} 
        & 96  & \textbf{\textcolor{red}{0.126}} & \textbf{\textcolor{red}{0.085}} & 0.428 & 0.271 & 0.395 & 0.268 & 0.346 & \textbf{\textcolor{blue}{0.234}} & 0.392 & 0.253 & 0.389 & 0.268 &  0.394 & 0.267 & 0.360 & 0.238 & \textbf{\textcolor{blue}{0.340}} & 0.240 & 0.375 & 0.251 & 0.455 & 0.298 & 0.481 & 0.280  \\

        & 192 & \textbf{\textcolor{red}{0.207}} & \textbf{\textcolor{red}{0.110}} & 0.448 & 0.282 & 0.417 & 0.276 & 0.371 & \textbf{\textcolor{blue}{0.246}}& 0.402 & 0.258 & 0.399 & 0.272 & 0.403 & 0.271 & 0.383 & 0.249 &  \textbf{\textcolor{blue}{0.343}} & 0.250 &  0.395 &  0.262 & 0.470 & 0.316 & 0.484 & 0.275  \\

        & 336 & \textbf{\textcolor{red}{0.279}} & \textbf{\textcolor{red}{0.182}} & 0.473 & 0.289 & 0.433 & 0.283 & 0.388 & \textbf{\textcolor{blue}{0.256}} & 0.428 & 0.263 & 0.417 & 0.279 & 0.417 & 0.278 & 0.395 & 0.259 & \textbf{\textcolor{blue}{0.363}} & 0.257 &  0.414 & 0.271 & 0.479 & 0.316 & 0.504 & 0.279  \\
        
        & 720 & \textbf{\textcolor{red}{0.373}} & \textbf{\textcolor{red}{0.268}} & 0.516 & 0.307 & 0.467 & 0.302 & 0.423 & 0.279 & 0.441 & 0.282 & 0.449 & 0.299 & 0.456 & 0.298 & 0.435 & 0.278 & \textbf{\textcolor{blue}{0.393}} & \textbf{\textcolor{blue}{0.271}} & 0.445 &  0.289 & 0.523 & 0.328 & 0.540 & 0.293  \\

        \midrule

        \multirow{5}{*}{\textbf{\rotatebox{90}{Weather}}} 
        & 96  & \textbf{\textcolor{red}{0.026}} & \textbf{\textcolor{red}{0.045}} & 0.157 & 0.205 & 0.174 & 0.214 & 0.154 & 0.196 & 0.155 & 0.205 & 0.174 & 0.231 &  0.146 & 0.198 & 0.146 & 0.191 & \textbf{\textcolor{blue}{0.144}} & \textbf{\textcolor{blue}{0.184}} & 0.153 &  0.199 & 0.156 & 0.204 & 0.168 & 0.203  \\

        & 192 & \textbf{\textcolor{red}{0.026}} & \textbf{\textcolor{red}{0.048}} & 0.204 & 0.247 & 0.221 & 0.254 & 0.201 & 0.240 & 0.201 & 0.245 & 0.216 & 0.267 & \textbf{\textcolor{blue}{0.185}} & 0.241 & 0.188 & 0.231 & 0.186 & \textbf{\textcolor{blue}{0.225}} & 0.202 & 0.246 & 0.209 & 0.249 & 0.214 & 0.245  \\

        & 336 & \textbf{\textcolor{red}{0.027}} & \textbf{\textcolor{red}{0.049}} & 0.261 & 0.290 & 0.278 & 0.296 &  0.257 & 0.283 & 0.237 & \textbf{\textcolor{blue}{0.265}} & 0.260 & 0.299 & 0.236 & 0.281 & \textbf{\textcolor{blue}{0.234}} & 0.268 & 0.237 & 0.267 & 0.260 & 0.289 & 0.264 & 0.290 & 0.236 & 0.273  \\
        
        & 720 & \textbf{\textcolor{red}{0.026}} & \textbf{\textcolor{red}{0.051}} & 0.340 & 0.341 & 0.358 & 0.347& 0.335 & 0.337 & 0.312 & 0.334 & 0.325 & 0.345 & 0.309 & 0.331 & \textbf{\textcolor{blue}{0.305}} & \textbf{\textcolor{blue}{0.319}} & 0.307 & 0.320 & 0.342 & 0.341 & 0.343 & 0.342 & 0.309 & 0.321  \\

        \midrule
        
        \multirow{5}{*}{\textbf{\rotatebox{90}{Exchange}}} 
        & 96  & \textbf{\textcolor{red}{0.004}} & \textbf{\textcolor{red}{0.038}} & - & - & 0.086 & 0.206 & 0.087 & 0.207 & 0.085 & 0.214 & 0.083 & 0.211 & 0.087 & 0.207 & \textbf{\textcolor{blue}{0.080}} & \textbf{\textcolor{blue}{0.198}} & - & - & - & - & 0.083 & 0.202 & 0.082 & 0.199  \\

        & 192 & \textbf{\textcolor{red}{0.008}} & \textbf{\textcolor{red}{0.056}} & - & - & 0.177 & 0.299 & 0.172 & 0.296 & 0.175 & 0.313 & 0.172 & 0.317 & 0.185 & 0.306 & \textbf{\textcolor{blue}{0.162}} & \textbf{\textcolor{blue}{0.288}} & - & - & - & - & 0.175 & 0.297 & 0.177 & 0.298  \\

        & 336 & \textbf{\textcolor{red}{0.014}} & \textbf{\textcolor{red}{0.078}} & - & - & 0.331 & 0.417 & 0.309 & 0.400 & 0.316 & 0.420 & 0.308 & 0.395 & 0.355 & 0.436 & \textbf{\textcolor{blue}{0.294}} & \textbf{\textcolor{blue}{0.392}} & - & - & - & - & 0.328 & 0.414 & 0.349 & 0.425  \\
        
        & 720 & \textbf{\textcolor{red}{0.034}} & \textbf{\textcolor{red}{0.126}} & - & - & 0.847 & 0.691 & 0.808 & 0.669 & 0.851 & 0.689 & 0.701 & 0.716 & 0.608 & 0.619 & \textbf{\textcolor{blue}{0.583}} & \textbf{\textcolor{blue}{0.580}} & - & - & - & - & 0.858 & 0.696 & 0.891 & 0.711  \\

        \midrule
        \textbf{1st Count}
        & & 24 & 32 & 0 & 0  & 0 & 0 & 1 & 0 & 0 & 0 & 0 & 0 & 2 & 0 & 3 & 0 & 3 & 0 & 0 & 0 & 0 & 0 & 0 & 0 \\

      \bottomrule
    \end{tabular}
  }
  \label{tab:main_table}
\end{table*}

\subsubsection{\textbf{Main Results}}
We evaluated our proposed pipeline for the multivariate long‐sequence time‐series forecasting (LSTF) task and report the results in Table \ref{tab:main_table}. In this table, the best results are marked in \textcolor{red}{red} and the second‐best results in \textcolor{blue}{blue}; lower mean squared error (MSE) and mean absolute error (MAE) values indicate superior forecasting accuracy. As shown in Table \ref{tab:main_table}, Our pipeline consistently outperforms all state-of-the-art methods across eight benchmark datasets.

The closest competitors, DUET, TimeBridge, GPHT, and TimeBase, remain unable to match its overall performance. Only on the ETT datasets does our method exhibit a few marginal failures, which we attribute to known inherent limitations of our foundation model, the iTransformer, which historically exhibits suboptimal performance on the ETT series. Despite the fatal failure of iTransformer on the ETT datasets, our pipeline’s ability to achieve top performance on ETT is particularly significant. Figure \ref{fig:result_plot} depicts the performance comparison between our proposed pipeline and existing state-of-the-art models for LSTF. The performance gains are substantial and clearly distinguishable from those of the other models. The robustness of our pipeline is further demonstrated across several foundational models, as reported in Table \ref{tab:more_models}.


\begin{table*}[h]
  \centering
  \caption{Performance comparison with alternative foundation models. iTransformer$\bigstar$ denotes our proposed pipeline with iTransformer as the foundation model. For other models, configurations without $\bigstar$ represent single-stage training baselines, while $\bigstar$ indicates integration as the foundation model in our pipeline.}
  \scriptsize
  \setlength{\tabcolsep}{2pt}
  \renewcommand{\arraystretch}{1.1}
  \resizebox{\textwidth}{!}{%
    \begin{tabular}{@{}c|c|cc|cc|cc|cc|cc|cc|cc|cc@{}}
      \toprule
      \multirow{2}{*}{\rotatebox{90}{\textbf{Dataset}}}
        & \multirow{2}{*}{\textbf{Seq\_len}}
        & \multicolumn{2}{c|}{\textbf{iTransformer}}
        & \multicolumn{2}{c|}{\textbf{iTransformer$\bigstar$}}
        & \multicolumn{2}{c|}{\textbf{GPHT}}
        & \multicolumn{2}{c|}{\textbf{TimeBridge}}
        & \multicolumn{2}{c|}{\textbf{TimeBase}}
        & \multicolumn{2}{c|}{\textbf{GPHT$\bigstar$}}
        & \multicolumn{2}{c|}{\textbf{TimeBridge$\bigstar$}}
        & \multicolumn{2}{c}{\textbf{TimeBase$\bigstar$}} \\
      \cmidrule(lr){3-4} \cmidrule(lr){5-6} \cmidrule(lr){7-8} \cmidrule(lr){9-10} \cmidrule(lr){11-12} \cmidrule(lr){13-14} \cmidrule(lr){15-16} \cmidrule(lr){17-18}
      & & \textbf{MSE} & \textbf{MAE} & \textbf{MSE} & \textbf{MAE} & \textbf{MSE} & \textbf{MAE} & \textbf{MSE} & \textbf{MAE} & \textbf{MSE} & \textbf{MAE} & \textbf{MSE} & \textbf{MAE} & \textbf{MSE} & \textbf{MAE} & \textbf{MSE} & \textbf{MAE} \\
      \midrule
      \multirow{4}{*}{\textbf{\rotatebox{90}{ETTh1}}}
        & 96  & 0.386 & 0.405 & 0.367 & 0.311 & 0.363 & 0.382 & 0.350 & 0.389 & 0.349 & 0.384 & 0.338 & 0.315 & 0.326 & 0.320 & 0.327 & 0.359 \\
        & 192 & 0.441 & 0.436 & 0.428 & 0.337 & 0.405 & 0.408 & 0.388 & 0.414 & 0.387 & 0.410 & 0.360 & 0.359 & 0.348 & 0.339 & 0.350 & 0.386 \\
        & 336 & 0.487 & 0.458 & 0.475 & 0.355 & 0.430 & 0.423 & 0.408 & 0.430 & 0.408 & 0.418 & 0.390 & 0.346 & 0.371 & 0.352 & 0.365 & 0.401 \\
        & 720 & 0.503 & 0.491 & 0.464 & 0.373 & 0.414 & 0.435 & 0.443 & 0.463 & 0.439 & 0.446 & 0.334 & 0.361 & 0.380 & 0.366 & 0.388 & 0.437 \\
      \midrule
      \multirow{4}{*}{\textbf{\rotatebox{90}{ETTm1}}}
        & 96  & 0.334 & 0.368 & 0.293 & 0.273 & 0.291 & 0.339 & 0.284 & 0.337 & 0.279 & 0.333 & 0.263 & 0.249 & 0.266 & 0.258 & 0.297 & 0.346 \\
        & 192 & 0.377 & 0.391 & 0.343 & 0.296 & 0.337 & 0.368 & 0.317 & 0.367 & 0.320 & 0.358 & 0.287 & 0.258 & 0.273 & 0.267 & 0.324 & 0.361 \\
        & 336 & 0.426 & 0.420 & 0.383 & 0.318 & 0.377 & 0.393 & 0.361 & 0.394 & 0.348 & 0.377 & 0.324 & 0.288 & 0.347 & 0.279 & 0.358 & 0.380 \\
        & 720 & 0.491 & 0.459 & 0.465 & 0.354 & 0.452 & 0.433 & 0.413 & 0.418 & 0.405 & 0.408 & 0.383 & 0.361 & 0.398 & 0.322 & 0.401 & 0.412 \\
      \midrule
      \multirow{4}{*}{\textbf{\rotatebox{90}{Weather}}}
        & 96  & 0.174 & 0.214 & 0.026 & 0.045 & 0.154 & 0.196 & 0.153 & 0.199 & 0.146 & 0.191 & 0.021 & 0.038 & 0.019 & 0.067 & 0.138 & 0.172 \\
        & 192 & 0.221 & 0.254 & 0.026 & 0.048 & 0.201 & 0.240 & 0.202 & 0.246 & 0.185 & 0.241 & 0.024 & 0.043 & 0.023 & 0.073 & 0.167 & 0.228 \\
        & 336 & 0.278 & 0.296 & 0.027 & 0.049 & 0.257 & 0.283 & 0.260 & 0.289 & 0.234 & 0.268 & 0.024 & 0.047 & 0.029 & 0.079 & 0.229 & 0.275 \\
        & 720 & 0.358 & 0.347 & 0.026 & 0.051 & 0.335 & 0.337 & 0.342 & 0.341 & 0.305 & 0.319 & 0.026 & 0.048 & 0.032 & 0.083 & 0.297 & 0.326 \\
      \midrule
      \multirow{4}{*}{\textbf{\rotatebox{90}{Traffic}}}
        & 96  & 0.395 & 0.268 & 0.126 & 0.085 & 0.346 & 0.234 & 0.375 & 0.251 & 0.360 & 0.238 & 0.109 & 0.072 & 0.132 & 0.106 & 0.385 & 0.260 \\
        
        & 192 & 0.417 & 0.276 & 0.207 & 0.110 & 0.371 & 0.246 & 0.395 & 0.262 & 0.383 & 0.249 & 0.182 & 0.093 & 0.214 & 0.126 & 0.392 & 0.266 \\
        
        & 336 & 0.433 & 0.283 & 0.279 & 0.182 & 0.388 & 0.256 & 0.414 & 0.271 & 0.395 & 0.259 & 0.256 & 0.161 & 0.263 & 0.195 & 0.409 & 0.267 \\
        
        & 720 & 0.467 & 0.302 & 0.373 & 0.268 & 0.423 & 0.279 & 0.445 & 0.289 & 0.435 & 0.278 & 0.354 & 0.257 & 0.360 & 0.274 & 0.449 & 0.291 \\
      \midrule
      \multirow{4}{*}{\textbf{\rotatebox{90}{Electricity}}}
        & 96  & 0.148 & 0.240 & 0.017 & 0.013 & 0.128 & 0.219 & 0.133 & 0.230 & 0.139 & 0.231 &0.013 &0.011 &0.015 &0.012 &0.110 & 0.189 \\
        
        & 192 & 0.162 & 0.253 & 0.023 & 0.014 & 0.146 & 0.236 & 0.154 & 0.248 & 0.153 & 0.245 & 0.018&0.012 &0.020 &0.016 &0.128 & 0.206 \\
        
        & 336 & 0.178 & 0.269 & 0.038 & 0.019 & 0.165 & 0.255 & 0.162 & 0.261 & 0.169 & 0.262 &0.027 &0.016 &0.041 &0.022 & 0.142 &0.239 \\
        
        & 720 & 0.225 & 0.317 & 0.045 & 0.026 & 0.207 & 0.292 & 0.184 & 0.284 & 0.207 & 0.294 &0.035 &0.023 &0.048 &0.028 &0.183 &0.255 \\
      \midrule
      \multirow{4}{*}{\textbf{\rotatebox{90}{Exchange}}}
        & 96  & 0.086 & 0.206 & 0.004 & 0.038 & 0.087 & 0.207 & - & - & 0.087 & 0.207 & 0.004 & 0.039 & - & - & 0.032 & 0.133 \\
        
        & 192 & 0.177 & 0.299 & 0.008 & 0.056 & 0.172 & 0.296 & - & - & 0.185 & 0.306 & 0.007 & 0.054 & - & - & 0.106 & 0.267 \\
        
        & 336 & 0.331 & 0.417 & 0.014 & 0.078 & 0.309 & 0.400 & - & - & 0.355 & 0.436 & 0.010 & 0.073 & - & - & 0.238 & 0.319 \\
        
        & 720 & 0.847 & 0.691 & 0.034 & 0.126 & 0.808 & 0.669 & - & - & 0.608 & 0.619 & 0.029 & 0.119 & - & - & 0.462 & 0.484 \\
      \bottomrule
    \end{tabular}
  }
  \label{tab:more_models}
\end{table*}


\begin{table}[h!]
\centering
\caption{Training and inference time overhead across datasets. While training requires approximately 2× time (one-time cost), inference maintains near-identical latency (1.00–1.006×).}
\footnotesize
\setlength{\tabcolsep}{3.5pt}
\begin{tabular}{@{}lccccc@{}}
\toprule
 & \textbf{ETTh1} & \textbf{ETTm2} & \textbf{Electricity} & \textbf{Weather} & \textbf{Traffic} \\
\midrule
\makecell[l]{${f}_{\theta}$ Training time (min)} &
1.67 & 4.76 & 18.25 & 8.98 & 19.66 \\
\addlinespace[0.2em]
\makecell[l]{${f}_{\phi}$ Training time (min)} &
1.43 & 4.82 & 19.06 & 6.22 & 19.48 \\
\addlinespace[0.2em]
\makecell[l]{Total Training Time (min)} &
3.10 & 5.98 & 37.31 & 15.20 & 39.14 \\
\addlinespace[0.2em]
\rowcolor{gray!20}
\makecell[l]{Training Overhead} &
1.86x & 2.00x & 2.00x & 1.69x & 1.99x \\
\midrule
\makecell[l]{${f}_{\theta}$ Inference (min)} &
0.53 & 2.39 & 3.16 & 2.63 & 5.09 \\
\addlinespace[0.2em]
\makecell[l]{${f}_{\phi}$ Inference (min)} &
0.53 & 2.40 & 3.18 & 2.64 & 5.11 \\
\addlinespace[0.2em]
\makecell[l]{Complete Inference Time (min)} &
0.53 & 2.40 & 3.18 & 2.64 & 5.11 \\
\addlinespace[0.2em]
\rowcolor{gray!20}
\makecell[l]{Inference Overhead} &
1.00x & 1.004x & 1.006x & 1.004x & 1.004x \\
\bottomrule
\end{tabular}
\label{tab:overhead}
\end{table}

\begin{table}[t!]
\centering
\caption{Model comparison: parameter count, performance ranking, inference speed, and computational complexity.}
\begin{tabular}{lcccl}
\hline
Model & Params & \makecell[l]{1st\\Count} & \makecell[l]{Inference\\Time} & Complexity \\
\hline
iTransformer & 5.47M & 0 & 2.743 ms/step & $O(N^2)$ \\
GPHT & 37.98M & 1 & 91.9 ms/step & $O(N^2 H)$ \\
TimeBridge & $\sim$5M & 3 & 2.56 ms/step & $O(L^2 M)$ \\
TimeBase & $\sim$0.5M & 2 & 0.98 ms/step & $O(aT + bL)$ \\
\rowcolor{gray!20}
Ours & 10.9M & \textbf{56} & 2.743 ms/step & $O(N^2)$ \\
\hline
\end{tabular}
\label{tab:parameter_complexity}
\end{table}

\subsubsection{\textbf{Performance Improvement}}

To quantify these improvements, we compare our pipeline against the three closest competitors, DUET, TimeBridge, and GPHT, across eight benchmark datasets in terms of MAE reduction. Compared to DUET, our pipeline improved performance by 17.83\% on ETTh1, 56.55\% on ETTh2, 16.06\% on ETTm1, 54.48\% on ETTm2, 92.85\% on Electricity, 79.94\% on Exchange, 38.36\% on Traffic, and 80.35\% on Weather. Similarly, compared to TimeBridge, it improved performance by 18.88\% on ETTh1, 56.41\% on ETTh2, 18.24\% on ETTm1, 54.31\% on ETTm2, 92.78\% on Electricity, 81.35\% on Exchange, 37.72\% on Traffic, and 79.98\% on Weather. Finally, compared to GPHT, our pipeline improved performance by 17.71\% on ETTh1, 88.30\% on ETTh2, 19.58\% on ETTm1, 55.91\% on ETTm2, 92.80\% on Electricity, 81.17\% on Exchange, 36.36\% on Traffic, and 81.82\% on Weather. These consistent, substantial improvements across diverse forecasting scenarios demonstrate the effectiveness and broad applicability of our approach.

\begin{figure*}[!t]
  \centering

  \begin{minipage}[t]{0.60\textwidth}
    \vspace{0pt}  
    \raggedright 
    \large
    \label{tab:ablation_final}
    \scriptsize
    \setlength{\tabcolsep}{4pt}
    \renewcommand{\arraystretch}{1.2}
    \resizebox{\linewidth}{!}{%
      \begin{tabular}{ccc|c|cc|cc|cc|cc|cc|cc}
        \toprule
        \multicolumn{3}{c|}{\textbf{Design}} &
        \multirow{2}{*}{\textbf{Horizon}} &
        \multicolumn{2}{c|}{\textbf{ETTh1}} &
        \multicolumn{2}{c|}{\textbf{ETTm1}} &
        \multicolumn{2}{c|}{\textbf{Traffic}} &
        \multicolumn{2}{c|}{\textbf{Weather}} &
        \multicolumn{2}{c|}{\textbf{Exchange}} &
        \multicolumn{2}{c}{\textbf{ECL}} \\
        \cmidrule(lr){1-3}
        \cmidrule(lr){5-6}
        \cmidrule(lr){7-8}
        \cmidrule(lr){9-10}
        \cmidrule(lr){11-12}
        \cmidrule(lr){13-14}
        \cmidrule(lr){15-16}
        $f_\theta$ & $f_\phi$ & $\alpha$ & & 
        \textbf{MSE} & \textbf{MAE} &
        \textbf{MSE} & \textbf{MAE} &
        \textbf{MSE} & \textbf{MAE} &
        \textbf{MSE} & \textbf{MAE} &
        \textbf{MSE} & \textbf{MAE} &
        \textbf{MSE} & \textbf{MAE} \\
        \midrule
        \multirow{4}{*}{$\checkmark$} & 
        \multirow{4}{*}{$\times$} & 
        \multirow{4}{*}{$\times$} & 96  & 
          0.386 & 0.405 & 0.334 & 0.368 & 0.395 & 0.268 & 0.174 & 0.214 & 0.086 & 0.206 & 0.148 & 0.240 \\
        & & & 192 & 
          0.441 & 0.436 & 0.377 & 0.391 & 0.417 & 0.276 & 0.221 & 0.254 & 0.177 & 0.299 & 0.162 & 0.253 \\
        & & & 336 & 
          0.487 & 0.458 & 0.426 & 0.420 & 0.433 & 0.283 & 0.278 & 0.296 & 0.331 & 0.417 & 0.178 & 0.269 \\
        & & & 720 & 
          0.503 & 0.491 & 0.491 & 0.459 & 0.467 & 0.302 & 0.358 & 0.347 & 0.847 & 0.691 & 0.225 & 0.317 \\
        \midrule
        \multirow{4}{*}{$\checkmark$} & 
        \multirow{4}{*}{$\checkmark$} & 
        \multirow{4}{*}{$\times$} & 96  & 
          0.400 & 0.335 & 0.342 & 0.308 & 0.295 & 0.170 & 0.026 & 0.050 & 0.006 & 0.049 & 0.114 & 0.183 \\
        & & & 192 & 
          0.438 & 0.352 & 0.366 & 0.320 & 0.367 & 0.224 & 0.026 & 0.051 & 0.010 & 0.065 & 0.138 & 0.206 \\
        & & & 336 & 
          0.484 & 0.359 & 0.401 & 0.337 & 0.398 & 0.265 & 0.028 & 0.052 & 0.017 & 0.087 & 0.169 & 0.233 \\
        & & & 720 & 
          0.475 & 0.379 & 0.480 & 0.369 & 0.428 & 0.303 & 0.027 & 0.053 & 0.037 & 0.132 & 0.180 & 0.248 \\
        \midrule
        \multirow{4}{*}{$\checkmark$} & 
        \multirow{4}{*}{$\checkmark$} & 
        \multirow{4}{*}{0.7} & 96  & 
          0.380 & 0.321 & 0.315 & 0.290 & 0.190 & 0.103 & 0.026 & 0.045 & 0.005 & 0.045 & 0.062 & 0.057 \\
        & & & 192 & 
          0.426 & 0.342 & 0.349 & 0.307 & 0.257 & 0.131 & 0.027 & 0.048 & 0.009 & 0.061 & 0.093 & 0.066 \\
        & & & 336 & 
          0.481 & 0.356 & 0.387 & 0.325 & 0.310 & 0.201 & 0.028 & 0.049 & 0.016 & 0.083 & 0.123 & 0.089 \\
        & & & 720 & 
          0.471 & 0.375 & 0.467 & 0.360 & 0.398 & 0.283 & 0.027 & 0.055 & 0.036 & 0.130 & 0.143 & 0.101 \\
        \midrule
        \multirow{4}{*}{$\checkmark$} & 
        \multirow{4}{*}{$\checkmark$} & 
        \multirow{4}{*}{\textbf{0.05}} & 96  & 
          \textbf{0.367} & \textbf{0.311} & 
          \textbf{0.292} & \textbf{0.272} & 
          \textbf{0.126} & \textbf{0.085} & 
          \textbf{0.025} & \textbf{0.045} & 
          \textbf{0.004} & \textbf{0.038} & 
          \textbf{0.017} & \textbf{0.013} \\
        & & & 192 & 
          \textbf{0.428} & \textbf{0.337} & 
          \textbf{0.338} & \textbf{0.294} & 
          \textbf{0.207} & \textbf{0.110} & 
          \textbf{0.026} & \textbf{0.047} & 
          \textbf{0.008} & \textbf{0.056} & 
          \textbf{0.023} & \textbf{0.014} \\
        & & & 336 & 
          \textbf{0.475} & \textbf{0.355} & 
          \textbf{0.377} & \textbf{0.316} & 
          \textbf{0.279} & \textbf{0.182} & 
          \textbf{0.027} & \textbf{0.049} & 
          \textbf{0.014} & \textbf{0.078} & 
          \textbf{0.038} & \textbf{0.019} \\
        & & & 720 & 
          \textbf{0.464} & \textbf{0.373} & 
          \textbf{0.459} & \textbf{0.351} & 
          \textbf{0.373} & \textbf{0.268} & 
          \textbf{0.026} & \textbf{0.051} & 
          \textbf{0.034} & \textbf{0.126} & 
          \textbf{0.045} & \textbf{0.026} \\
        \bottomrule
      \end{tabular}%
    }
    \captionof{table}{Ablation Results on our proposed pipeline. To facilitate a direct comparison with our two‐stage pipeline, we compare results obtained with the base model $f_\theta$ alone versus those obtained with both the base $f_\theta$ and meta model $f_\phi$, and evaluate the impact of $\alpha$ by comparing “no $\alpha$” against different values (0.7 vs.\ our selected 0.05).}
    \label{tab:ablation_main_paper}
  \end{minipage}
  \hfill
  \begin{minipage}[t]{0.39\textwidth}
    \vspace{2pt}
    \centering
    \includegraphics[width=1.1\linewidth]{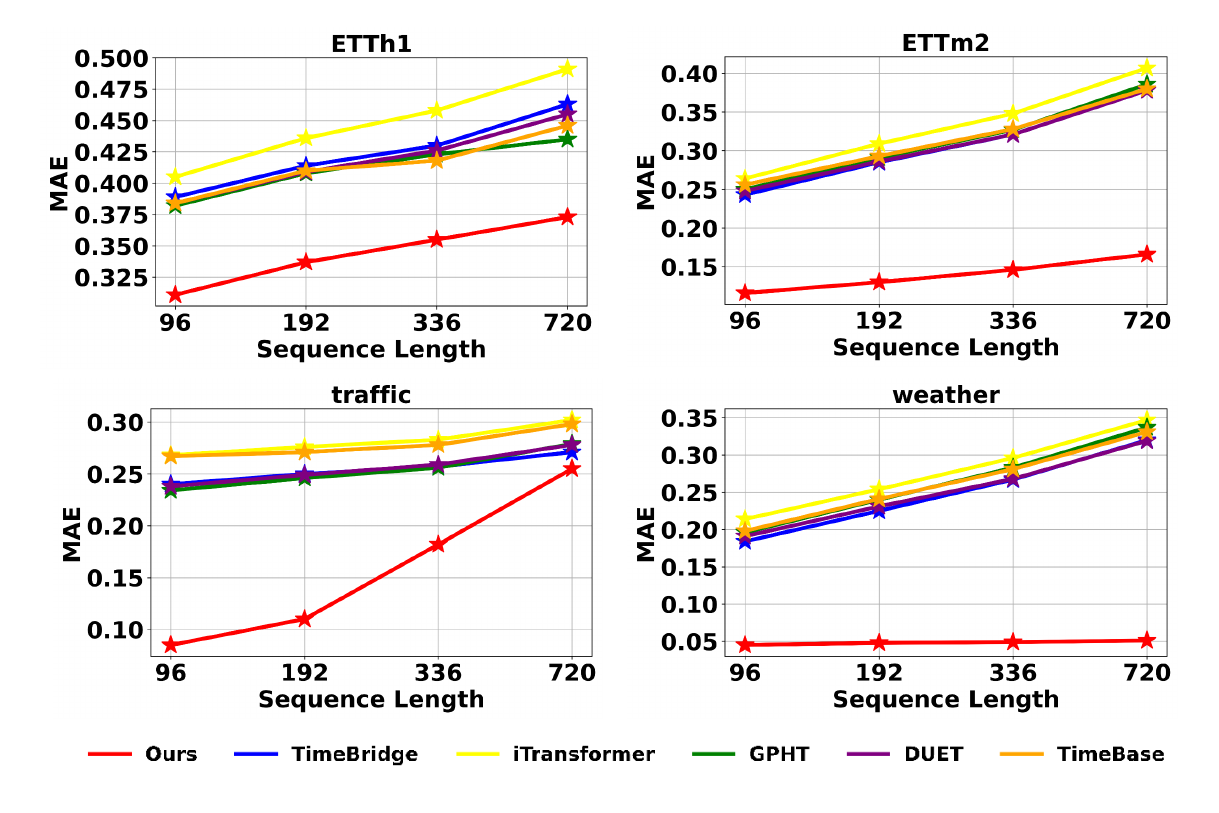}
    \captionsetup{width=\linewidth}
    \caption{Forecasting performance evaluation with prediction horizons $S \in \{96,192,336,720\}$ and a fixed lookback length of 96. The proposed residual-aware learning framework improves accuracy across both short and long horizons.}
    \label{fig:result_plot}
  \end{minipage}

\end{figure*}

\subsubsection{\textbf{Robustness Across Foundation Models}} We evaluated our pipeline with alternative foundational models, including transformer-based architectures (GPHT \cite{liu2024generative} and TimeBridge \cite{liu2024timebridge}) and an MLP-based model (TimeBase \cite{huangtimebase}). Replacing iTransformer with GPHT yielded gains of up to 15.79\%; however, this doubles the parameter count of GPHT (37.98M × 2 = 75.96M), resulting in a 7× increase, leading us to reject GPHT due to scalability concerns in resource-constrained environments. TimeBridge and TimeBase also showed improvements, confirming our pipeline's model-agnostic nature, though neither consistently outperformed our iTransformer-based pipeline across all benchmarks. Details in Table \ref{tab:more_models}. TimeBridge showed comparable but not superior gains to iTransformer except on the ETT datasets, and TimeBase yielded only modest improvements. Our pipeline consistently improves LSTF performance across foundational models, with the largest gains from transformer-based architectures; iTransformer offers the best balance between accuracy and efficiency. 

The key findings are summarized below:

\textbf{Model Agnostic:} Our proposed pipeline improves LSTF performance irrespective of the foundational model selection, demonstrating broad applicability across diverse architectures.
\textbf{Model Selection:}  iTransformer represents an optimal choice when balancing performance improvements against model size and resource-constrained environments.

\subsubsection{\textbf{Pipeline Complexity and Computational Overhead}} Training $f_{\theta}$ and $f_{\phi}$ sequentially doubles parameters and training time versus single-stage models, but maintains $O(N_1^2 + N_2^2) \approx O(N^2)$ complexity, matching iTransformer. During inference, both models operate independently in parallel, resulting in negligible overhead: the meta-model $f_{\phi}$ requires only the initial window computed from the base model $f_{\theta}$'s predictions and ground truths, after which both models execute concurrently. Despite these doubled training costs, our 10.9M-parameter pipeline remains significantly lighter than existing baselines such as GPHT (37.98M parameters) while achieving substantial performance gains up to 92.85\% improvement and 56 first-place rankings across all benchmark configurations. This combination of competitive parameter efficiency and superior performance justifies the additional training overhead, which represents a one-time cost that does not impact deployment. Detailed breakdowns are provided in Table~\ref{tab:overhead} and Table~\ref{tab:parameter_complexity}.

\subsubsection{\textbf{ Ablation Study}}

To validate the design rationale of our proposed pipeline's components, we conduct detailed ablation studies that include experiments involving the replacement and removal of specific components. The corresponding results are reported in Table \ref{tab:ablation_main_paper}. We evaluated performance by (1) removing the scaling factor applied to the residual predictions of the meta model $f_\phi$, and (2) using alternative value of $\alpha$. In both cases, our proposed pipeline employing the chosen $\alpha$ value outperformed these alternative configurations. This finding underscores the consistent superiority of our proposed multi stage pipeline over alternative design paradigms across benchmark datasets.


\subsection{Discussion}

Our results set a new state-of-the-art for Long-term Sequence Time-series Forecasting (LSTF), outperforming existing methods. The two closest competitors, DUET and TimeBridge, both use transformer architectures, validating our architectural choice and confirming transformers' effectiveness for temporal LSTF modeling. However, our pipeline exhibited suboptimal performance on certain instances within the ETT dataset, which can be attributed to inherent limitations of our foundational iTransformer architecture. This phenomenon aligns with documented catastrophic failures of iTransformer on ETT datasets reported in prior literature, suggesting that dataset-specific architectural considerations may be necessary for optimal performance across diverse temporal domains. A critical distinction in our evaluation is the use of autoregressive testing protocols. Unlike previous approaches that use ground truth values as inputs during evaluation, our models operate under realistic conditions where previous predictions serve as inputs for subsequent forecasts. While this may cause recursive error accumulation, it provides a more authentic assessment of real-world performance where future ground truths are unavailable. Theoretical foundation, empirical validation, and residual correction visualization demonstrate strong alignment between theory and observations. The results indicate that the residual correction mechanism behaves as intended, with empirical patterns closely matching theoretical expectations.

\section{Conclusion}
This paper presents a novel two-stage residual-aware representation learning pipeline that addresses fundamental limitations in single-stage transformer-based time series forecasting by explicitly decoupling forecasting and residual correction into specialized stages. The methodology treats model-generated residuals as structured, learnable signals rather than random noise, introducing a meta-corrector model that dynamically captures error patterns across multivariate channels while preserving cross-variable dependencies. Comprehensive evaluation across eight benchmark datasets demonstrates substantial performance improvements, achieving state-of-the-art results with gains up to 92.85\% compared to leading methods including DUET, TimeBridge, and GPHT. The framework's model-agnostic design applies to both transformer-based and alternative architectures. Its success in mitigating systematic biases suggests that explicit residual modeling represents a meaningful advancement toward accurate long-term forecasting, opening new research directions in multi-stage predictive architectures.


\bibliographystyle{ACM-Reference-Format}
\bibliography{sample-base}

\clearpage
\appendix
\renewcommand{\appendixname}{Supplementary}

\section{Theoretical Foundation and Empirical Validation of Hypothesis Space Expansion}
\label{sec:Theoretical_Foundation}

\noindent\textbf{Central Claim:} The two-stage pip expands the learnable hypothesis space beyond single-stage models by enabling specialized learning of structured residual patterns, thereby reducing approximation error.

\subsection{Formal Framework}

\noindent\textit{Hypothesis Space Expansion.} Let $\mathcal{H}_{\text{base}}$ and $\mathcal{H}_{\text{meta}}$ denote the hypothesis classes of the base model and meta-corrector, respectively. The two-stage prediction is given by
\begin{equation}
\label{eq:two_stage_prediction}
\hat{y} = f_{\theta}(X) - \alpha \cdot f_{\phi}(\mathcal{R}),
\end{equation}
where $f_{\theta} \in \mathcal{H}_{\text{base}}$ represents the base forecasting model, $f_{\phi} \in \mathcal{H}_{\text{meta}}$ represents the meta-corrector, $\mathcal{R} = f_{\theta}(X) - y$ denotes the residual series (prediction error), and $\alpha \in [0,1]$ is a scaling factor.

Let $f^*_{\theta}$ denote the optimal base forecasting model obtained through single-stage training on the original series $X$:
\begin{equation}
\label{eq:optimal_base}
f^*_{\theta} = \arg\min_{f_{\theta} \in \mathcal{H}_{\text{base}}} \|y - f_{\theta}(X)\|^2,
\end{equation}
where $y$ denotes the ground truth and $\|\cdot\|^2$ represents the mean squared error (MSE). We define $\mathcal{R} = f^*_{\theta}(X) - y$ as the residual series produced by the optimal base model after single-stage training.

If the two-stage pipeline reduces the LSFT forecasting error, then the following condition must hold:

\begin{equation}
\label{eq:hypothesis_space}
\|y - (f^*_{\theta}(X) - \alpha \cdot f_{\phi}(\mathcal{R}))\|^2 < \|y - f^*_{\theta}(X)\|^2.
\end{equation}

\noindent\textit{\textbf{Proof sketch.}} The two-stage prediction error can be decomposed as follows:
\begin{align*}
\|y - (f^*_{\theta}(X) - \alpha \cdot f_{\phi}(\mathcal{R}))\|^2 &= \|y - f^*_{\theta}(X) + \alpha \cdot f_{\phi}(\mathcal{R})\|^2 \\
&= \|-\mathcal{R} + \alpha \cdot f_{\phi}(\mathcal{R})\|^2 \\
&= \|\mathcal{R} - \alpha \cdot f_{\phi}(\mathcal{R})\|^2.
\end{align*}
From Eq.~\eqref{eq:hypothesis_space}, the two-stage error reduction condition becomes:
\begin{equation}
\label{eq:error_reduction}
\|\mathcal{R} - \alpha \cdot f_{\phi}(\mathcal{R})\|^2 < \|\mathcal{R}\|^2.
\end{equation}

\noindent\textbf{Theoretical Proposition.} The error reduction condition through hypothesis space expansion in Eq.~\eqref{eq:error_reduction} holds if and only if the meta-model $f_{\phi}$ can accurately learn and predict the residual series $\mathcal{R}$, with appropriate scaling by $\alpha$. The scaling factor $\alpha$ serves as a regularization hyperparameter that assists the meta-model $f_{\phi}$ in preventing overcorrection. Applying $\alpha$ to erroneous residual predictions will not yield error reduction through hypothesis expansion. These theoretical considerations lead to two specific, testable predictions:
\begin{itemize}
    \item \textbf{Prediction 1:} Residuals from the base model contain systematic, learnable patterns rather than white noise.
    \item \textbf{Prediction 2:} The meta-corrector's performance depends critically on the preservation of residual structure; destroying this structure should eliminate the performance gain.
\end{itemize}

\noindent\textbf{Null Hypothesis.} If hypothesis space expansion is invalid, then the residual series $\mathcal{R}$ contains no learnable patterns beyond white noise, and random residual predictions scaled by $\alpha$ would equivalently minimize the two-stage prediction error.

\subsection{Empirical Validation}

\subsubsection{\textbf{Test 1:} Evidence for Structured Residuals (Prediction 1)}

\noindent\textbf{Method.} We apply two independent statistical tests to assess the structure of residual series across all datasets:
\begin{enumerate}
    \item \textbf{Ljung-Box Q-test:} Tests the null hypothesis that residuals constitute white noise with no autocorrelation.
    \item \textbf{Autocorrelation analysis:} Examines the persistence of temporal dependencies across different lag intervals.
\end{enumerate}

\noindent\textbf{Results. }
The Ljung-Box test in yields $p < 10^{-6}$ across all residual datasets, strongly rejecting the white noise hypothesis at any conventional significance level. The full table is provided in Table \ref{tab:ljung}. Autocorrelation functions (Figure~\ref{fig:autocorrelation}) reveal significant autocorrelation persisting across multiple lags, consistently exceeding the 95\% confidence bands. More details in \ref{sub:Evidence for Structured Residuals}.

\subsubsection{Test 2: Preservation of Residual Structure and Existence of Learnable Patterns (Prediction 2)}

If the error reduction arises from learning structured residual patterns rather than random chance, then performance should collapse when residual structure is destroyed.

\noindent\textbf{Method.} We conduct counterfactual experiments that systematically destroy residual structure:
\begin{enumerate}
    \item \textbf{Time-shuffled residuals:} Randomly permute residual time indices, destroying temporal dependencies while preserving the marginal distribution.
    \item \textbf{Gaussian noise replacement:} Replace residuals with i.i.d. Gaussian noise matched to the empirical mean and variance.
\end{enumerate}

\noindent\textbf{Results.}
We train the meta-model using Huber loss on both shuffled and randomly generated residual series. First, we evaluate the meta-model $f_{\phi}$'s predictive performance on these corrupted series in Table~\ref{tab:shuffle_random}. The meta-model can predict the shuffled residual series with performance comparable to that achieved on the actual sequential series with intact temporal relations, but fails dramatically on the random series, yielding poor performance.

However, when integrating the predictions from meta-models trained on shuffled or random residual series into our proposed two-stage pipeline, performance degrades in both cases rather than improving. Complete results are presented in Table~\ref{tab:final_result_shuffled_random}.

\subsection{Concluding Statement}

Having empirically validated both predictions through rigorous statistical tests and counterfactual experiments, we reject the null hypothesis that residual series constitute unstructured white noise and that random residual predictions contribute to error reduction. The results conclusively demonstrate that the two-stage pipeline's performance gains arise from a principled expansion of the hypothesis space through structured residual learning, rather than from spurious correlations or statistical artifacts.


\section{Limitations of Traditional Boosting/Hybrid Designs and Our Contributions}
\label{sec:improve_residual_model}

Our proposed pipeline provides several capabilities that traditional boosting and hybrid approaches lack for long-sequence time forecasting (LSTF), and therefore addresses key deficiencies that make those classic designs ill-suited for this setting. Our main contributions are listed below:

\noindent\textbf{Representational Improvements}

\begin{itemize}
    \item \textbf{Limitation:} Boosting/hybrid approaches impose ad-hoc decompositions (linear vs nonlinear) or rely on residuals being white noise; they do not explicitly expand the hypothesis class in a formal way.
    \item \textbf{Our fix:} formalize residual learning as a hypothesis-space expansion: base predictor $f\theta$ earns primary dynamics; meta-corrector $f_\phi$
    adds complementary functions over residual space so the combined family can represent functions that a single model cannot. This directly addresses the representational gap and epistemic approximation error that single models or heuristic hybrids leave behind.
\end{itemize}

\noindent\textbf{Parallel Multi-Horizon Inference.}
\begin{itemize}
  \item \textbf{Limitation:} Hybrid setups rely on one-step or recursive residual correction (or require sequentially applying a residual predictor per horizon), causing significant error accumulation and linear-in-horizon inference cost.
  \item \textbf{Our fix:} Both base and meta output the entire forecasting horizon in a single forward pass and run in parallel, removing recursion, reducing accumulated error, and making inference horizon-cost constant per model pass.
\end{itemize}

\noindent\textbf{Scale-Separated Residual Correction.}
\begin{itemize}
  \item \textbf{Limitation:} Hybrid models often train the residual predictor on raw predictions/residuals on the same scale as the base; mismatched transforms or unnormalized residuals allow the meta model to overshoot or learn inefficiently.
  \item \textbf{Our fix:} We train the meta model on the actual scale of the residual series so the meta model targets structured residual components on the correct scale, improving stability and interpretability.
\end{itemize}

\noindent\textbf{Statistical stability, training protocol and data leakage}
\begin{itemize}
  \item \textbf{Limitation:} Hybrid / stacking pipelines can leak information or overfit when training and residual computation aren’t strictly isolated. Boosting ensembles may overfit when lag features are excessive.
  \item \textbf{Our fix:} We adpot a two-stage training scheme with frozen base model weights (no shared parameters) and strictly isolate datasets for the meta model so that residual learning is not contaminated by leakage. This force the meta model to learn true residual structure and improving generalization and training stability.
\end{itemize}

\noindent\textbf{Targeted capacity \& hyperparameter limitations (boosting/hybrid)}
\begin{itemize}
\item \textbf{Limitation:} Boosting/hybrid methods force a single global capacity/hyperparameter design onto both coarse and fine signals, leading to oversized ensembles, brittle tuning, and inefficient parameter use.
\item \textbf{Our fix:} Our two-stage pipeline separates duties — allocate capacity, losses and optimizers per stage (e.g., choose different base/meta sizes; distinct loss function and optimizers), increasing flexibility, effective capacity, and training stability.
\end{itemize}

\section{Pipeline}

\subsection{Foundational model selecting criterias:}
\label{sub:Foundational model selecting}

We select iTransformer as our foundational model for both base model $f_{\theta}$ and meta model $f_{\phi}$ based on few criterias such as- 

Transformer-based models consistently outperform linear and CNN baselines on nonlinear dynamics \cite{qiu2024tfb}. Linear models are confined to linear spaces and treat variates independently, while CNNs impose locality biases through convolutions, both failing to capture complex temporal patterns and distant interactions in residuals.

iTransformer maintains the closest fidelity to the vanilla Transformer architecture while adapting SOTA performance for LSTF balancing parameter vs performance tradeoff. A near-vanilla transformer isolates core inductive biases (self-attention, multi-head, positional encoding), ensuring observed behaviors are attributable to the architecture rather than auxiliary modifications. Using other transformer based models such as GPHT, TimeXer, DUET, or TimeBridge, which introduce additional components or impose heavy modifications to the vanilla transformer architecture raise the question wheather the performance improvement is the reason of our proposed pipeline or due to the modified sophisticated architecture or component. While our main goal is to establish the explicit effectiveness of our pipeline, without getting influenced by any edditional component or architectural modifications, we select iTransformer at the first place. 

To justify the selection of iTransformer as our foundational model, we conducted additional experiments replacing it with alternative transformer-based and non-transformer architectures. Comprehensive performance comparisons are presented in Appendix Table \ref{tab:more_models}, while model complexity and computational overhead analyses are detailed in Tables \ref{tab:parameter_complexity} and \ref{tab:overhead}, respectively.

\subsection{Addressing Overfitting and Data Leakage}
\label{sub-sec:Addressing Overfitting and Data Leakage}
A vigorous challenge of this pipeline is to ensure that the training protocol is preserved and crucial subjects such as overfitting and data leakage are countered. Overfitting leads to memorizing the training data and failing to generalize on unseen data. To address this issue we trained two models for two distinct learning objectives. First, a base model $f_{\theta}$ learns the actual pattern of the original dataset features. On the other hand, the second meta correction model $f_{\phi}$ learns the possible amount residual for the base model $f_{\theta}$. To prevent the data leakage, both models are trained on completely isolated datasets and no model weights or dataset statistical properties are shared among the models. While the base model $f_{\theta}$ is trained using the original dataset, the meta correction model $f_{\phi}$  remains idle and no dataset information or weights are shared to the meta model $f_{\phi}$. After finishing the training of the base model $f_{\theta}$ and generating the base prediction $\mathcal{P}$, training of the base model $f_{\theta}$ is terminated and the residual series $\mathcal{R}$ is obtained. The meta model $f_{\phi}$ is then trained on the residual series $\mathcal{R}$ without providing prior knowledge about the weights of the previous base models $f_{\theta}$ or the relation of the original data set with the residual data set. While training the meta model $f_{\phi}$ , the base model $f_{\theta}$  remains completely idle and even after finishing the training process of the meta model $f_{\phi}$, no further training or weight update is done on the base model $f_{\theta}$.


\subsection{Sign Altering in Residual Series }
\label{sub-sec:Sign Altering in Residual Series}

In the original dataset, the sign of a given feature remains relatively stable throughout the entire time series. However, in the fluctuation dataset, the sign of each feature varies frequently over time, resulting in rapid transitions between positive and negative values. These frequent sign changes obscure inherent patterns, making them more challenging to discern. In some instances, this volatility increases the risk of the model capturing white noise rather than meaningful fluctuations.

This introduces a significant challenge for the model, as it must accurately predict both the sign and magnitude of fluctuations. Since these predicted fluctuations are incorporated into the final calculation, errors in either sign or magnitude can lead to a substantial degradation in overall performance, potentially producing results worse than those obtained with the original model.

To mitigate this issue, we replaced the Mean Squared Error (MSE) loss of iTransformer \cite{liu2023itransformer} with the Huber loss. Huber loss offers a more robust approach by effectively handling both sign and magnitude errors, making it particularly advantageous in this context. Consequently, adopting the Huber loss can enhance performance in predicting fluctuation values compared to MSE.

\begin{equation}
    Hubber Loss_{\delta}(a) =
\begin{cases} 
\frac{1}{2} a^2, & \text{if } |a| \leq \delta \\
\delta (|a| - \frac{1}{2} \delta), & \text{if } |a| > \delta
\end{cases}
\end{equation}

Where, \(\alpha= \hat y- y \) represents the difference between the predicted and true values and $\delta$ is a hyperparameter that controls the threshold for the error term, determining the transition point between the quadratic and linear regions of the loss function. When there's a sign mismatch between predicted and actual fluctuations, this represents a significant error since predicting $+x$ when the true value is $-x$ means the correction will worsen rather than improve the base prediction. In this case, the absolute error $|a|$ will be larger than $\delta$, putting us in the linear regime of the Huber loss: $\delta(|a| - ½\delta)$. This linear scaling for large errors is crucial for sign prediction in two ways: First, unlike MSE which would impose an overwhelming quadratic penalty (potentially causing gradient explosions or making the model too conservative), the linear penalty provides a strong but manageable gradient signal that helps the model learn to correct sign errors without becoming unstable. Second, the transition from quadratic to linear at $\delta$ creates a clear "boundary" in the loss landscape that the model can learn from - errors beyond this threshold (which often correspond to sign mismatches) are treated distinctly from smaller errors. This helps the model develop a better "awareness" of sign importance while still maintaining sensitivity to magnitude through the quadratic regime for smaller errors. Additionally, because the linear growth prevents extreme penalties for large errors, the model can more freely explore and learn sign patterns during training without being overly penalized for occasional sign mistakes, leading to better generalization in sign prediction.

\subsection{Preventing Over-correction of Meta Model}
\label{sub:Preventing Over-correction of Meta Model}
While formulating the final prediction combining both base model $f_{\theta}$ and meta-correction model $f_{\phi}$, employing subtraction to calibrate the base predictions introduces an additional challenge. In a two‐stage forecasting framework, the meta‐corrector must predict both the correct sign and magnitude of the base‐model residuals: if residual prediction ${\mathbf{\mathcal{E}}}$ has the wrong sign or a large wrong magnitude, its calibration can degrade rather than improve the forecast. Only when the majority of residual predictions align with the true error in both sign and magnitude does the cascaded pipeline reliably reduce MSE and MAE; conversely, a few large erroneous residuals can negate many small correct ones, potentially worsening performance relative to the base model.

Applying a constant scaling factor $\alpha \in [0,1]$ to the meta-model’s $f_{\phi}$ residual predictions ${\mathbf{\mathcal{E}}}$ effectively attenuates the corrective adjustments; large or sign‑mismatched residual predictions are softened, thereby mitigating the risk of error overshoot. This is analogous to using a smaller learning rate in boosting or gradient descent, which stabilizes each update and tames variance. Seminal studies confirm this effect: Friedman \cite{friedman2001greedy} showed that using $\eta<1$ (shrinkage) in gradient boosting dramatically reduced test error, and Chen and Guestrin \cite{chen2016xgboost} explicitly scale each tree’s output by $\eta<1$ for the same reason.

\noindent\textbf{Selecting the scaling factor $\alpha$:} 
The scaling factor $\alpha$ is a non-learnable hyperparameter applied to mittigate over-correction by ${f_{\phi}}$. To identify a robust value that generalizes across datasets, we perform a GridSearch over $\alpha \in [0, 1]$ with step size 0.05, evaluating all eight benchmark datasets. For each candidate value, we compute the final MSE and MAE across all dataset-horizon combinations (8 datasets × 4 horizons = 32 settings). We select $\alpha = 0.05$ as it achieves the highest number of best performances (wins) across these 32 configurations, demonstrating consistent effectiveness in mitigating over-correction without requiring dataset-specific tuning. We select $\alpha$ = 0.05 as it achieves the highest number of best performances across these 32 configurations

\noindent\textbf{Distinction from Other Residual Scaling Approaches:} We note that residual scaling is employed in various deep learning contexts, though with fundamentally different motivations. Methods like feature normalization \cite{alshaher2021studying}, gradient clipping \cite{huang2024improved}, and $\mu$P \cite{dey2024sparse} scale residuals to address training dynamics, improving gradient flow, preventing instability, or enabling hyperparameter transfer during model optimization. In contrast, our scaling factor $\alpha$ addresses an inference-time prediction problem: preventing over-correction when the meta-corrector's residual forecasts have incorrect sign or magnitude. This positions our work closer to shrinkage in gradient boosting \cite{friedman2001greedy, chen2016xgboost}, where learning rate scaling dampens individual learners' contributions to improve ensemble robustness and prevent overfitting.

\section{Experiment Details}
\label{sec:experiment_details_appendix}
\subsection{Dataset Details}
\label{sub-sec:Dataset Details append}

We evaluate the efficacy of our proposed pipeline on eight real‑world long‑term forecasting benchmarks, highlighting four representative datasets: 
Electricity Transformer Temperature (ETT) , containing four subsets (ETTh1/ETTh2 at hourly intervals and ETTm1/ETTm2 at 15‑minute intervals) of transformer oil temperature readings, which we forecast as the endogenous series using six power‑load features as exogenous inputs;

Electricity (ECL) , which comprises hourly consumption records from 321 clients—where we treat the last client’s series as the endogenous target and the remaining 320 series as exogenous inputs;

Traffic, featuring hourly road‑occupancy rates from 862 sensors on San Francisco Bay Area freeways, where we predict the last sensor’s occupancy as the endogenous series and employ the other 861 sensors’ measurements as exogenous variables.

Weather , consisting of 21 meteorological measurements taken every 10 minutes at the Max Planck Institute for Biogeochemistry in 2020, with wet‑bulb temperature as the endogenous variable and the other 20 indicators as exogenous covariates;

and, Exchange comprises a panel of daily foreign exchange rates for eight countries, spanning the period from 1990 to 2016.

The details of all datasets are listed in Table \ref{tab:complete_datasets_stat}.

\begin{table}[!htbp]
    \centering
    \caption{Overview of datasets used in our experiments}
    \label{tab:complete_datasets_stat}
    \scriptsize
    \setlength{\tabcolsep}{6pt}
    \renewcommand{\arraystretch}{1.2}
    \begin{tabular}{@{}l c c c c@{}}
        \toprule
        Dataset     & Features & prediction Length        & Dataset Size            & Frequency \\ 
        \midrule
        ETTm1       & 7   & \{96, 192, 336, 720\} & \{34465, 11521, 11521\}  & 15 min    \\
        ETTm2       & 7   & \{96, 192, 336, 720\} & \{34465, 11521, 11521\}  & 15 min    \\
        ETTh1       & 7   & \{96, 192, 336, 720\} & \{ 8545,  2881,  2881\}  & 15 min    \\
        ETTh2       & 7   & \{96, 192, 336, 720\} & \{ 8545,  2881,  2881\}  & 15 min    \\
        Electricity & 321 & \{96, 192, 336, 720\} & \{18317,  2633,  5261\}  & Hourly    \\
        Traffic     & 862 & \{96, 192, 336, 720\} & \{12185,  1757,  3509\}  & Hourly    \\
        Exchange    & 8   & \{96, 192, 336, 720\} & \{ 5120,   665,  1422\}  & Daily     \\
        Weather     & 21  & \{96, 192, 336, 720\} & \{36792,  5271, 10540\}  & 10 min    \\
        \bottomrule
    \end{tabular}
\end{table}

\subsection{Data Processing}
We follow the data preprocessing and chronological train–validation–test splitting protocol of iTransformer (built on TimesNet), thereby preventing any data leakage. For the ETT, Weather, ECL, Solar‑Energy, and Traffic datasets, we use a fixed look‑back window of 96 time steps and evaluate forecasting horizons of 96, 192, 336, and 720.

\subsection{Hyperparameter Settings}
\label{sub-sec:Hyperparameter Settings append}

During experimentation, we performed systematic hyperparameter tuning to evaluate the influence of each parameter on forecasting performance. All models were trained for a maximum of 10 epochs with early stopping (patience = 3) to prevent overfitting. To ensure computational efficiency, we benchmarked performance across multiple GPU platforms including NVIDIA A5000, A6000, RTX 3090, and RTX 4090, and selected the RTX 4090 based on its superior throughput. While several hyperparameters were kept consistent across all datasets to maintain experimental rigor, others were adapted to the specific characteristics of individual datasets. We first summarize the key hyperparameters in Table~\ref{tab:hyperparameter_dataset}, followed by a detailed explanation of their selected values.

\noindent\textbf{Notations for the hyperparameters:}

\begin{description}[
    noitemsep,
    labelwidth=\widthof{\bfseries seq\_len},
    labelsep=1em,
    leftmargin=!  
  ]
  \item[\textbf{e\_layers}] Number of stacked encoder layers (model depth).
  \item[\textbf{d\_model}] Dimensionality of the Transformer’s hidden embeddings.
  \item[\textbf{d\_ff}] Hidden‑layer size in the position‑wise feed‑forward network.
  \item[\textbf{n\_heads}] Number of attention heads in each multi‑head block.
  \item[\textbf{embed}] Timestamp encoding method:
    \begin{itemize}[noitemsep,topsep=0pt]
      \item \texttt{timeF}: sinusoidal calendar features
      \item \texttt{fixed}: classic positional embeddings
      \item \texttt{learned}: learnable embedding vectors
    \end{itemize}
\end{description}

\begin{table}[!htbp]
    \centering
    \caption{Dataset dependent hyperparameters}
    \scriptsize
    \setlength{\tabcolsep}{4pt}
    \renewcommand{\arraystretch}{1.2}
    \adjustbox{max width=\columnwidth}{ 
    \begin{tabular}{c|c c c c c}
        \toprule
        \textbf{Dataset} & \textbf{e\_layers} & \textbf{d\_model} & \textbf{d\_ff} & \textbf{batch size} & \textbf{learning\_rate} \\
        
        \midrule
        
        \textbf{ETTh1}
        & 2  & 256 & 256 & 32 & 0.0001 \\

        \midrule
        \textbf{ETTh2}
        & 2 & 128 & 128 & 32 & 0.0001 \\

        \midrule
        \textbf{ETTm1}
        & 2 & 128 & 128 & 32 & 0.0001 \\

        \midrule
        \textbf{ETTm2}
        & 2 & 128 & 128 & 32 & 0.0001 \\

        \midrule
        \textbf{ECL}
        & 3 & 512 & 512 & 16 & 0.0005 \\

        \midrule
        \textbf{Traffic}
        & 4 & 512 & 512 & 16 & 0.001 \\

        \midrule
        \textbf{Weather}
        & 3 & 512 & 512 & 32 & 0.0001 \\

        \midrule
        \textbf{Exchange}
        & 2 & 128 & 128 & 32 & 0.0001 \\
        
        \bottomrule
    \end{tabular}
    }
    \label{tab:hyperparameter_dataset}
\end{table}

\noindent\textbf{Additional general hyperparameters:}

\begin{description}[
    noitemsep,
    labelwidth=\widthof{\bfseries seq\_len},
    labelsep=1em,
    leftmargin=!  
  ]
  \item[\textbf{dropout}] 0.1
  \item[\textbf{embed}] timeF
  \item[\textbf{activation}] gelu
  \item[\textbf{num\_workers}] 10 
  \item[\textbf{optimizer}] Adam
  \item[\textbf{n\_heads}] 8
\end{description}

\section{Additional Results}

\subsection{Complete Results of Residual‑Series Prediction}

Table \ref{tab:full_fluctuation_performance} reports the comprehensive residual‑series forecasting performance across all eight datasets, evaluated using both the Huber loss and mean‑squared error objectives. Forecasted values have been inverse‑transformed to their original scale, enabling error metrics to be expressed in real‑world units. Presenting results on this scale enhances interpretability of magnitude improvements, thereby facilitating practical assessment and comparison for real‑world applications.

\begin{table}[!htbp]
    \centering
    \caption{Comprehensive Evaluation of Residual Series Prediction Using MSE and Huber Loss}
    \scriptsize
    \setlength{\tabcolsep}{4pt}
    \renewcommand{\arraystretch}{1.2}
    \adjustbox{max width=\columnwidth}{ 
    \begin{tabular}{c|c|cc|cc}
        \toprule
        \multirow{2}{*}{\textbf{\rotatebox{90}{Dataset}}} & \multirow{2}{*}{\textbf{Metric}} & \multicolumn{2}{c|}{\textbf{MSE Loss}} & \multicolumn{2}{c}{\textbf{Huber loss}} \\
        & & \textbf{MSE} & \textbf{MAE} & \textbf{MSE} & \textbf{MAE} \\
        \midrule
        \multirow{4}{*}{\textbf{\rotatebox{90}{ETTh1}}} 
        & 96  & 9.820 & 1.697 & 9.804 & 1.693 \\
        & 192 & 12.425 & 1.935 & 12.466 & 1.936 \\
        & 336 & 13.847 & 2.078 & 13.987 & 2.077 \\
        & 720 & 12.598 & 2.030 & 12.754 & 2.035 \\

        \midrule
        \multirow{5}{*}{\textbf{\rotatebox{90}{ETTh2}}} 
        & 96  & 29.761 & 3.458 & 29.555 & 3.436 \\
        & 192 & 41.796 & 4.212 & 41.382 & 4.184 \\
        & 336 & 52.911 & 4.798 & 52.082 & 4.756 \\
        & 720 & 46.439 & 4.579 & 46.002 & 4.554 \\

        \midrule
        \multirow{5}{*}{\textbf{\rotatebox{90}{ETTm1}}} 
        & 96  & 8.652 & 1.598 & 8.595 & 1.583 \\
        & 192 & 10.742 & 1.798 & 10.722 & 1.791 \\
        & 336 & 12.373 & 1.949 & 12.337 & 1.939 \\
        & 720 & 10.740 & 1.699 & 10.689 & 1.689 \\

        \midrule
        \multirow{5}{*}{\textbf{\rotatebox{90}{ETTm2}}} 
        & 96  & 13.018 & 2.194 & 12.399 & 2.117 \\
        & 192 & 14.719 & 2.321 & 14.202 & 2.261 \\
        & 336 & 15.225 & 2.372 & 14.745 & 2.316 \\
        & 720 & 15.118 & 2.392 & 14.697 & 2.343 \\

        \midrule
        \multirow{5}{*}{\textbf{\rotatebox{90}{Electricity}}} 
        & 96  & 7162381.5 & 282.279 & 6941128.500 & 279.050 \\
        & 192 & 10383775.0 & 320.369 & 10151179.0 & 315.395 \\
        & 336 & 12942538.0 & 355.415 & 12024151.0 & 345.789 \\
        & 720 & 14548618.2 & 386.590 & 14297862.7 & 380.568 \\

        \midrule
        \multirow{5}{*}{\textbf{\rotatebox{90}{Traffic}}} 
        & 96  & 0.000859 & 0.014976 & 0.000859 & 0.014716 \\
        & 192 & 0.0009157 & 0.015434 & 0.0009200 & 0.015221 \\
        & 336 & 0.001256 & 0.018165 & 00.001026 & 0.01689 \\
        & 720 & 0.001768 & 0.020812 & 0.001543 & 0.01936 \\

        \midrule
        \multirow{5}{*}{\textbf{\rotatebox{90}{Weather}}} 
        & 96  & 3597.358 & 20.161 & 3398.848 & 19.616\\
        & 192 & 3715.890 & 21.244 & 3665.362 & 20.906 \\
        & 336 & 3703.213 & 21.983 & 3639.305 & 21.853 \\
        & 720 & 3856.499 & 23.430 & 3867.301 & 23.507 \\

        \midrule
        \multirow{5}{*}{\textbf{\rotatebox{90}{Exchange}}} 
        & 96  & 0.000584 & 0.015531 & 0.000572 & 0.015312 \\
        & 192 & 0.000737 & 0.017713 & 0.000727 & 0.017569 \\
        & 336 & 0.000846 & 0.019310 & 0.000835 & 0.019176 \\
        & 720 & 0.0008002 & 0.018962 & 0.000794 & 0.018886 \\

        \bottomrule
    \end{tabular}
    }
    \label{tab:full_fluctuation_performance}
\end{table}

\subsection{ More Baselines in Long-term Forecasting}

To ensure a fair and comprehensive evaluation, we include additional baselines in Table \ref{tab:main_table_two}. Over the past several years, diverse architectural paradigms have demonstrated strong performance on the long‑term time‑series forecasting (LSTF) task. Accordingly, we have incorporated the most widely adopted and high‑performing models from recent literature to rigorously benchmark the efficacy of our proposed pipeline. Except for a few isolated cases on the ETT dataset, our results consistently outperform all baseline models.

\subsection{Comprehensive Robustness Analysis}
\label{sub:robustness foundational models}

We replaced our foundational model iTransformer with other state-of-the-art models, including transformer-based architectures (GPHT \cite{liu2024generative} and TimeBridge \cite{liu2024timebridge}) and an MLP-based model (TimeBase \cite{huangtimebase}).

Existing benchmarks show that GPHT outperforms iTransformer on conventional LSTF tasks with single-stage training, and we observe a similar pattern here. Replacing iTransformer with GPHT yielded performance gains of up to 15.79\%. This improvement likely stems from GPHT's inherent architectural design, whereas iTransformer represents a minimally modified vanilla transformer. However, using GPHT as the foundational model doubles the parameter count (37.98M × 2 = 75.96M), resulting in approximately 7× more parameters than the iTransformer-based pipeline (10.9M). While the performance improvements validate our pipeline's effectiveness, achieving only a 15.79\% gain at the cost of a 7-fold parameter increase led us to reject GPHT as the foundational model due to scalability concerns in resource-constrained environments. Similarly, replacing iTransformer with TimeBridge and TimeBase yielded performance improvements, clearly demonstrating our pipeline's robustness across different model selections. However, several observations warrant careful attention. In single-stage training on the eight benchmark datasets, both TimeBridge and TimeBase outperform iTransformer in most cases. Yet when integrated into our pipeline, TimeBridge's performance gains remain comparable to, but not superior to, the iTransformer-based pipeline. Meanwhile, deploying TimeBase within our pipeline results in only modest performance gains. Specifically, the TimeBridge-integrated pipeline occasionally outperforms the iTransformer-based pipeline, except on the ETT dataset. Conversely, the TimeBase-integrated pipeline fails to outperform the iTransformer-based pipeline in all cases except the ETT dataset.

The key findings are summarized below:
\begin{itemize}
\item \textbf{Model Agnostic:} Our proposed pipeline improves LSTF performance irrespective of the foundational model selection, demonstrating broad applicability across diverse architectures.
\item \textbf{Model Selection:} Transformer-based models tend to outperform other architectures in LSTF performance gains within our pipeline. iTransformer represents an optimal choice when balancing performance improvements against model size and resource-constrained environments.
\end{itemize}

\subsection{Performance Analysis on Special Cases}
\label{sub:special case results}

Single-stage models struggle with overlapping multi-scale patterns—trends, seasonal cycles, and high-frequency variations—because their unified optimization prioritizes dominant features, leaving complex overlapping dynamics as unmodeled residuals. Our two-stage pipeline introduces an additional learning paradigm specifically designed to capture these leftover patterns: after the base model learns primary temporal structures, the meta-corrector explicitly models the systematic errors arising from unrepresented overlapping patterns, preventing any structured component from being ignored. ADF tests confirm strong performance on non-stationary data (e.g., Exchange, $p=0.41$). Despite suboptimal results on the stationary ETTh1 dataset ($p=0.0083$), attributable to inherent architectural limitations of iTransformer, our pipeline generalizes effectively across both stationary and non-stationary series. Furthermore, following TFB~\cite{qiu2024tfb} Algorithm 1, we computed distribution shift indicators $\delta$ for all datasets, where $\delta > 0.8$ defines significant distribution shift. Our pipeline maintains robust performance despite substantial shifts in ETTh2, ETTm2, Traffic, and Weather datasets, demonstrating effectiveness in handling sudden changes in data distribution (complete $\delta$ values in Appendix Table~\ref{tab:delta_values}).

\begin{table}[!htbp]
    \centering
    \caption{Distribution shift indicators ($\delta$) for each dataset. $\delta > 0.8$ indicates significant distribution shift.}
    \scriptsize
    \setlength{\tabcolsep}{4pt}
    \renewcommand{\arraystretch}{1.2}
    \noindent\adjustbox{max width=0.48\textwidth,left}{
    \begin{tabular}{cccccccc}
        \toprule
        \textbf{ETTh1} & \textbf{ETTh2} & \textbf{ETTm1} & \textbf{ETTm2} & \textbf{exchange} & \textbf{traffic} & \textbf{weather} & \textbf{electricity} \\
        \midrule
        0.17 & 0.97 & 0.17 & 0.91 & 0.79 & 0.93 & 0.99 & 0.53 \\
        \bottomrule
    \end{tabular}
    }
    \label{tab:delta_values}
\end{table}


\subsection{Evidence for Structured Residuals and Preservation Requirements}
\label{sub:Evidence for Structured Residuals}

\noindent\textbf{White Noise Test.} We conducted the Ljung-Box Q-test on all residual series across multiple lag values (7, 10, 20, 30), with results presented in Table~\ref{tab:ljung}. The test yields $p < 10^{-6}$ across all datasets, strongly rejecting the white noise hypothesis and confirming the presence of structured patterns in the residuals.

Additionally, autocorrelation analysis (Figure~\ref{fig:autocorrelation}) reveals high positive autocorrelation at lag 1 with slow, gradual decay over subsequent lags. All early lags significantly exceed the 95\% confidence bands, with persistent patterns observable through lag 40. These characteristics collectively demonstrate that the residual series exhibits strong autocorrelation structure and is definitively not white noise.

\begin{table}[!htbp]
    \centering
    \caption{Ljung-Box Q-test P values for the eight benchmark datasets}
    \scriptsize
    \setlength{\tabcolsep}{4pt}
    \renewcommand{\arraystretch}{1.2}
    \noindent\adjustbox{max width=0.48\textwidth,left}{
    \begin{tabular}{cccccccc}
        \toprule
        \textbf{ETTh1} & \textbf{ETTh2} & \textbf{ETTm1} & \textbf{ETTm2} & \textbf{exchange} & \textbf{traffic} & \textbf{weather} & \textbf{electricity} \\
        \midrule
        0.0000 & 0.0000 & 0.0000 & 0.0000 & 0.0000 & 0.0000 & 0.0000 & 0.0000 \\
        \bottomrule
    \end{tabular}
    }
    \label{tab:ljung}
\end{table}

\noindent\textbf{Preservation of Residual Structure.} To validate that performance gains arise from learning structured patterns rather than statistical artifacts, we conducted counterfactual experiments by destroying residual structure through: (1) time-shuffling, which randomizes temporal order while preserving the marginal distribution, and (2) Gaussian noise replacement with matched variance. We trained the meta-model $f_{\phi}$ on both corrupted residual series and evaluated its predictive performance (Table~\ref{tab:shuffle_random}). The meta-model achieves competitive performance on shuffled residuals—comparable to training on sequential residuals, but fails catastrophically on random noise, yielding poor predictions. Critically, when integrating predictions from meta-models trained on corrupted residuals into our two-stage pipeline, performance degrades in both cases rather than improving (Table~\ref{tab:final_result_shuffled_random}). The shuffled-residual meta-model's predictions, despite being competitively accurate to original sequential residual series, fail to reduce base model error when applied in the pipeline, while random-residual predictions actively increase error. This demonstrates that hypothesis space expansion is not merely a random error correction mechanism. Effective residual correction requires more than accurate residual prediction in isolation, it depends critically on preserving the structured temporal relationship between residuals and base model predictions.

\begin{table}[!htbp]
    \centering
    \caption{Predictive performance of meta-model $f_{\phi}$ trained on corrupted residuals using Huber loss. Results show MSE and MAE across datasets when training on time-shuffled residuals (temporal order destroyed) and random Gaussian noise (all structure destroyed)}
    \scriptsize
    \setlength{\tabcolsep}{4pt}
    \renewcommand{\arraystretch}{1.2}
    \adjustbox{max width=\columnwidth}{
    \begin{tabular}{c|c|cc|cc}
        \toprule
        \multirow{2}{*}{\textbf{\rotatebox{90}{Dataset}}} & \multirow{2}{*}{\textbf{Metric}} & \multicolumn{2}{c|}{\textbf{Shuffled Series}} & \multicolumn{2}{c}{\textbf{Random Series}} \\
        \cmidrule(lr){3-4} \cmidrule(lr){5-6}
        & & \textbf{MSE} & \textbf{MAE} & \textbf{MSE} & \textbf{MAE} \\
        \midrule
        \multirow{4}{*}{\textbf{\rotatebox{90}{ETTh1}}} 
        & 96  & 10.43 & 1.88 & 26.62 & 4.32 \\
        & 192 & 14.54 & 2.04 & 34.42 & 11.89 \\
        & 336 & 11.65 & 2.56 & 41.52 & 16.60 \\
        & 720 & 15.76 & 2.66 & 63.13 & 22.28 \\
        \midrule
        \multirow{4}{*}{\textbf{\rotatebox{90}{ETTh2}}} 
        & 96  & 24.57 & 4.01 & 46.15 & 7.95 \\
        & 192 & 51.65 & 4.16 & 80.31 & 10.51 \\
        & 336 & 46.51 & 5.32 & 112.64 & 16.73 \\
        & 720 & 52.90 & 4.88 & 136.39 & 15.32 \\
        \midrule
        \multirow{4}{*}{\textbf{\rotatebox{90}{ETTm1}}} 
        & 96  & 9.11 & 1.89 & 14.37 & 1.583 \\
        & 192 & 13.43 & 2.17 & 17.93 & 1.791 \\
        & 336 & 10.62 & 2.98 & 28.80 & 1.939 \\
        & 720 & 16.75 & 2.74 & 41.46 & 1.689 \\
        \midrule
        \multirow{4}{*}{\textbf{\rotatebox{90}{ETTm2}}} 
        & 96  & 11.83 & 2.52 & 27.13 & 5.17 \\
        & 192 & 16.67 & 2.90 & 38.35 & 11.29 \\
        & 336 & 14.78 & 3.22 & 51.54 & 19.38 \\
        & 720 & 17.12 & 3.05 & 46.41 & 28.61 \\
        \midrule
        \multirow{4}{*}{\textbf{\rotatebox{90}{Electricity}}} 
        & 96  & 7026462.34 & 283.124 & 7813151.66 & 366.518 \\
        & 192 & 11371611.12 & 340.455 & 12425561.72 & 512.415 \\
        & 336 & 12648612.58 & 369.764 & 15468756.91 & 720.115 \\
        & 720 & 15642064.23 & 401.457 & 18662351.63 & 906.363 \\
        \midrule
        \multirow{4}{*}{\textbf{\rotatebox{90}{Traffic}}} 
        & 96  & 0.000976 & 0.016256 & 0.008026 & 0.065494 \\
        & 192 & 0.001020 & 0.018346 & 0.011365 & 0.135685 \\
        & 336 & 0.001256 & 0.020189 & 0.023698 & 0.182611 \\
        & 720 & 0.001713 & 0.022561 & 0.027164 & 0.265655 \\
        \midrule
        \multirow{4}{*}{\textbf{\rotatebox{90}{Weather}}} 
        & 96  & 3516.46 & 21.18 & 5120.424 & 33.512 \\
        & 192 & 3765.54 & 23.54 & 5468.549 & 45.452 \\
        & 336 & 3620.17 & 22.77 & 6022.420 & 53.564 \\
        & 720 & 3792.57 & 26.65 & 6896.346 & 75.844 \\
        \midrule
        \multirow{4}{*}{\textbf{\rotatebox{90}{Exchange}}} 
        & 96  & 0.000601 & 0.017165 & 0.00893 & 0.091354 \\
        & 192 & 0.000755 & 0.018313 & 0.00123 & 0.168549 \\
        & 336 & 0.000964 & 0.019315 & 0.00161 & 0.315648 \\
        & 720 & 0.000892 & 0.026465 & 0.00236 & 0.365484 \\
        \bottomrule
    \end{tabular}
    }
    \label{tab:shuffle_random}
\end{table}


\begin{table}[h]
  \centering
  \caption{Two-stage pipeline performance with meta-models trained on sequential, shuffled, and random residuals. "Ours" uses $f_{\phi}$ trained on sequential residuals; "Shuffled" and "Random" use $f_{\phi}$ trained on time-shuffled and noise residuals, respectively. Base model $f_{\theta}$ predictions remain constant all setups. Both corrupted configurations degrade performance.}
  \scriptsize
  \setlength{\tabcolsep}{2.5pt}
  \renewcommand{\arraystretch}{1.1}
  \resizebox{\columnwidth}{!}{%
    \begin{tabular}{@{}c|c|cc|cc|cc|cc@{}}
      \toprule
      \multirow{2}{*}{Dataset}
        & \multirow{2}{*}{\makecell[l]{Models\\Metric}}
        & \multicolumn{2}{c}{\makebox[2.5pt][c]{Ours}}
        & \multicolumn{2}{c}{\makebox[2.5pt][c]{iTransformer}}
        & \multicolumn{2}{c}{\makebox[2.5pt][c]{Shuffled}}
        & \multicolumn{2}{c}{\makebox[2.5pt][c]{Random}} \\
      \cmidrule(lr){3-4} \cmidrule(lr){5-6} \cmidrule(lr){7-8} \cmidrule(lr){9-10}
      & & MSE & MAE & MSE & MAE & MSE & MAE & MSE & MAE \\
      \midrule

      \multirow{4}{*}{ETTh1} 
        & 96  & 0.367 & 0.311 & 0.386 & 0.405 & 0.397 & 0.413 & 0.621 & 0.843 \\
        & 192 & 0.428 & 0.337 & 0.441 & 0.436 & 0.472 & 0.491 & 0.765 & 0.697 \\
        & 336 & 0.475 & 0.355 & 0.487 & 0.458 & 0.504 & 0.517 & 0.723 & 0.618 \\
        & 720 & 0.464 & 0.373 & 0.503 & 0.491 & 0.521 & 0.511 & 0.910 & 0.813 \\
      \midrule

      \multirow{4}{*}{ETTh2} 
        & 96  & 0.054 & 0.144 & 0.297 & 0.349 & 0.358 & 0.379 & 0.671 & 0.782 \\
        & 192 & 0.066 & 0.161 & 0.380 & 0.400 & 0.397 & 0.411 & 0.924 & 0.631 \\
        & 336 & 0.078 & 0.178 & 0.428 & 0.432 & 0.442 & 0.450 & 0.661 & 0.885 \\
        & 720 & 0.078 & 0.183 & 0.427 & 0.445 & 0.468 & 0.513 & 0.780 & 0.921 \\
      \midrule

      \multirow{4}{*}{ETTm1}
        & 96  & 0.293 & 0.273 & 0.334 & 0.368 & 0.378 & 0.391 & 0.667 & 0.834 \\
        & 192 & 0.343 & 0.296 & 0.377 & 0.391 & 0.399 & 0.422 & 0.712 & 0.653 \\
        & 336 & 0.383 & 0.318 & 0.426 & 0.420 & 0.502 & 0.488 & 0.916 & 0.807 \\
        & 720 & 0.465 & 0.354 & 0.491 & 0.459 & 0.518 & 0.501 & 0.788 & 0.980 \\
      \midrule

      \multirow{4}{*}{ETTm2} 
        & 96  & 0.037 & 0.116 & 0.180 & 0.264 & 0.212 & 0.306 & 0.632 & 0.549 \\
        & 192 & 0.048 & 0.130 & 0.250 & 0.309 & 0.287 & 0.321 & 0.591 & 0.872 \\
        & 336 & 0.059 & 0.146 & 0.311 & 0.348 & 0.377 & 0.410 & 0.638 & 0.710 \\
        & 720 & 0.074 & 0.166 & 0.412 & 0.407 & 0.467 & 0.455 & 0.882 & 0.743 \\
      \midrule

      \multirow{4}{*}{Electricity} 
        & 96  & 0.017 & 0.013 & 0.148 & 0.240 & 0.223 & 0.371 & 0.765 & 0.564 \\
        & 192 & 0.023 & 0.014 & 0.162 & 0.253 & 0.234 & 0.330 & 0.763 & 0.672 \\
        & 336 & 0.038 & 0.019 & 0.178 & 0.269 & 0.343 & 0.428 & 0.879 & 0.787 \\
        & 720 & 0.045 & 0.026 & 0.225 & 0.317 & 0.367 & 0.411 & 0.567 & 0.879 \\
      \midrule

      \multirow{4}{*}{Traffic} 
        & 96  & 0.126 & 0.085 & 0.395 & 0.268 & 0.444 & 0.371 & 0.675 & 0.569 \\
        & 192 & 0.207 & 0.110 & 0.417 & 0.276 & 0.526 & 0.467 & 0.688 & 0.865 \\
        & 336 & 0.279 & 0.182 & 0.433 & 0.283 & 0.576 & 0.398 & 0.659 & 0.889 \\
        & 720 & 0.373 & 0.268 & 0.467 & 0.302 & 0.572 & 0.466 & 0.658 & 0.748 \\
      \midrule

      \multirow{4}{*}{Weather} 
        & 96  & 0.026 & 0.045 & 0.174 & 0.214 & 0.254 & 0.297 & 0.589 & 0.982 \\
        & 192 & 0.026 & 0.048 & 0.221 & 0.254 & 0.304 & 0.285 & 0.656 & 0.689 \\
        & 336 & 0.027 & 0.049 & 0.278 & 0.296 & 0.346 & 0.387 & 0.652 & 0.854 \\
        & 720 & 0.026 & 0.051 & 0.358 & 0.347 & 0.479 & 0.462 & 0.854 & 0.758 \\
      \midrule

      \multirow{4}{*}{Exchange} 
        & 96  & 0.004 & 0.038 & 0.086 & 0.206 & 0.124 & 0.344 & 0.858 & 0.782 \\
        & 192 & 0.008 & 0.056 & 0.177 & 0.299 & 0.327 & 0.406 & 0.654 & 0.586 \\
        & 336 & 0.014 & 0.078 & 0.331 & 0.417 & 0.481 & 0.711 & 0.686 & 0.887 \\
        & 720 & 0.034 & 0.126 & 0.847 & 0.691 & 0.976 & 0.819 & 0.895 & 0.768 \\
      \bottomrule
    \end{tabular}
  }
  \label{tab:final_result_shuffled_random}
\end{table}

\section{Scalability}
\label{sec:Scalability}

\subsection{Dataset and Sequence Scaling}
Our pipeline demonstrates robust scalability across diverse dataset sizes (8,545 to 18,317 training samples), prediction horizons (96, 192, 336, 720 time steps with fixed 96-step lookback), and high-dimensional scenarios (up to 862 features in Traffic, 321 in Electricity). Performance remains consistently superior to baselines across all scales.

\subsection{Training and Inference Efficiency}
Training time exhibits predictable scaling behavior, ranging from 3.10 minutes (ETTh1: 8,545 samples, 7 features) to 39.14 minutes (Traffic: 12,185 samples, 862 features) for 10 epochs with early stopping on a single NVIDIA RTX 4090 GPU (Table \ref{tab:overhead}). Critically, inference maintains near-identical latency to the base model with negligible overhead (1.00×–1.006×) through parallel execution of $f_{\theta}$ and $f_{\phi}$. After constructing the initial residual input, both models process subsequent windows simultaneously using autoregressively predicted residuals, ensuring real-time forecasting feasibility even for large-scale datasets (Traffic: 5.11 min, Electricity: 3.18 min for complete test evaluation).

\noindent\textbf{GPU Utilization Efficiency}

Despite operating two models, our pipeline maintains GPU utilization nearly identical to the single-stage baseline (Table~\ref{tab:gpu_util}). This efficient resource utilization validates that the two-stage architecture introduces negligible computational overhead during deployment, ensuring practical scalability for production environments.

\begin{table}[!htbp]
\centering
\small
\caption{GPU utilization metrics on NVIDIA RTX 4090. Our pipeline achieves near identitical efficiency to the single-stage baseline.}
\label{tab:gpu_util}
\begin{tabular}{lcccc}
\toprule
\textbf{Model} & \textbf{Avg} & \textbf{Peak} & \textbf{Mem BW} & \textbf{FLOPs} \\
 & \textbf{GPU} & \textbf{GPU} & \textbf{(GB/s)} & \textbf{Eff.} \\
\midrule
iTransformer & 80.6\% & 98.4\% & 612 & 82.1\% \\
Our Pipeline & 78.8\% & 96.5\% & 598 & 81.4\% \\
\bottomrule
\end{tabular}
\end{table}

\subsection{Parameter Efficiency}
With 10.9M parameters (5.47M each for $f_{\theta}$ and $f_{\phi}$), our pipeline achieves up to 92.85\% MAE improvement while using 3.5× fewer parameters than GPHT (37.98M). Across 32 experimental settings (8 datasets × 4 horizons), we secured 56 first-place rankings (Table 2), demonstrating that doubling parameters yields order-of-magnitude accuracy gains. Sequential training ensures no memory overhead during training, and parallel inference maintains efficient resource utilization.


\section{Additional Ablation Studies}

As shown before, the Huber loss performs better than MSE loss while predicting the residual series in two stage prediction pipeline. We further experimented with this Huber loss function by directly replacing the the original MSE loss function of iTransformer with the Huber loss and run the model in a single stage to predict the future values. The findings are listed in Table \ref{tab:ablation_two}. 


\begin{table}[!htbp]
    \centering
    \caption{Additional ablation results on our proposed pipeline. To facilitate a direct comparison with our two-stage pipeline, we substituted the mean squared error (MSE) loss of the single-stage iTransformer with the Huber loss for the original time-series forecasting task.}
    \scriptsize
    \setlength{\tabcolsep}{4pt}
    \renewcommand{\arraystretch}{1.2}
    \noindent\adjustbox{max width=0.48\textwidth,left}{
    \begin{tabular}{c|c|c|cc|cc|cc|cc|cc|cc}
        \toprule
        \multirow{2}{*}{\textbf{\rotatebox{90}{Design}}} & 
        \multirow{2}{*}{\textbf{\rotatebox{90}{Loss}}} & 
        \multirow{2}{*}{\makecell[l]{\textbf{Dataset$\rightarrow$} \\\textbf{Metric$\rightarrow$}}} & 
        \multicolumn{2}{c|}{\textbf{ETTh1}} & 
        \multicolumn{2}{c}{\textbf{ETTm1}} & 
        \multicolumn{2}{c}{\textbf{Traffic}} & 
        \multicolumn{2}{c}{\textbf{Weather}} & 
        \multicolumn{2}{c}{\textbf{Exchange}} & 
        \multicolumn{2}{c}{\textbf{ECL}} \\
        
        & & &
        \textbf{MSE} & \textbf{MAE} & 
        \textbf{MSE} & \textbf{MAE} & 
        \textbf{MSE} & \textbf{MAE} & 
        \textbf{MSE} & \textbf{MAE} & 
        \textbf{MSE} & \textbf{MAE} & 
        \textbf{MSE} & \textbf{MAE} \\
        \midrule
        \multirow{4}{*}{\textbf{\rotatebox{90}{our pipeline}}} 
        & \multirow{4}{*}{\textbf{\rotatebox{90}{MSE ,Huber}}} 
        
        & 96  & \textbf{0.367} & \textbf{0.311} & \textbf{0.293} & \textbf{0.273} & \textbf{0.126} & \textbf{0.085} & \textbf{0.026} & \textbf{0.045} & \textbf{0.004} & \textbf{0.038} & \textbf{0.017} & \textbf{0.013} \\
        
        & & 192 & \textbf{0.428} & \textbf{0.337} & \textbf{0.343} & \textbf{0.296} & \textbf{0.207} & \textbf{0.110} & \textbf{0.026} & \textbf{0.048} & \textbf{0.008} & \textbf{0.056} & \textbf{0.023} & \textbf{0.014} \\
        
        & & 336 & \textbf{0.475} & \textbf{0.355} & \textbf{0.383} & \textbf{0.318} & \textbf{0.279} & \textbf{0.182} & \textbf{0.027} & \textbf{0.049} & \textbf{0.014} & \textbf{0.078} & \textbf{0.038} & \textbf{0.019} \\
        
        & & 720 & \textbf{0.464} & \textbf{0.373} & \textbf{0.465} & \textbf{0.354} & \textbf{0.373} & \textbf{0.268} & \textbf{0.026} & \textbf{0.051} & \textbf{0.034} & \textbf{0.126} & \textbf{0.045} & \textbf{0.026} \\

        \midrule
        \multirow{4}{*}{\rotatebox{90}{iTransformer}} 
        & \multirow{4}{*}{\rotatebox{90}{MSE}}
        
        & 96  & 0.386 & 0.405 & 0.334 & 0.368  & 0.395 & 0.268 & 0.174 & 0.214 & 0.086 & 0.206 & 0.148 & 0.240 \\
        
        & & 192 & 0.441 & 0.436 & 0.377 & 0.391 & 0.417 & 0.276 & 0.221 & 0.254 & 0.177 & 0.299 & 0.162 & 0.253 \\
        
        & & 336 & 0.487 & 0.458 & 0.426 & 0.420 & 0.433 & 0.283 & 0.278 & 0.296 & 0.331 & 0.417 & 0.178 & 0.269 \\
        
        & & 720 & 0.503 & 0.491 & 0.491 & 0.459 & 0.467 & 0.302 & 0.358 & 0.347 & 0.847 & 0.691 & 0.225 & 0.317 \\
        
        \midrule
        \multirow{4}{*}{\rotatebox{90}{iTransformer}} 
        & \multirow{4}{*}{\rotatebox{90}{Huber}} 
        & 96  & 0.387 & 0.400 & 0.335 & 0.362 & 0.396 & 0.280 & 0.178 & 0.217 & 0.089 & 0.210 & 0.153 & 0.243 \\
        & & 192 & 0.438 & 0.429 & 0.381 & 0.386 & 0.433 & 0.284 & 0.225 & 0.257 & 0.183 & 0.304 & 0.168 & 0.256 \\
        & & 336 & 0.470 & 0.444 & 0.416 & 0.410 & 0.451 & 0.294 & 0.283 & 0.299 & 0.348 & 0.429 & 0.185 & 0.273 \\
        & & 720 & 0.474 & 0.466 & 0.482 & 0.447 & 0.482 & 0.309 & 0.360 & 0.349 & 0.866 & 0.702 & 0.224 & 0.305 \\
        
        \bottomrule
    \end{tabular}
    }
    \label{tab:ablation_two}
\end{table}

\section{Error Bars}
We conducted experiments with both fixed and varying random seeds to evaluate result consistency and robustness. Each setting was repeated three times. Error analysis with same seed is listed in Table \ref{tab:error_bar_same_seed} and with different  seeds listed in Table \ref{tab:error_bar_different_seed_actual_result} and \ref{tab:error_bar_different_seed_stats_confidence}.

\begin{table}[!htbp]
    \centering
    \caption{Repeating all the experiments three times with the same seed and reporting MSE, MAE and confidence level}
    \scriptsize
    \setlength{\tabcolsep}{4pt}
    \renewcommand{\arraystretch}{1.2}
    \adjustbox{max width=\columnwidth}{%
    \begin{tabular}{c|c|cccc}
        \toprule
        \multirow{2}{*}{\rotatebox{90}{\textbf{Dataset}}} &
        \multirow{2}{*}{\textbf{Horizon}} &
        \multicolumn{2}{c}{\textbf{Metrics}} & \multicolumn{2}{c}{\textbf{Confidence (\%)}} \\
        \cmidrule(lr){3-6}
        & & \textbf{MSE} & \textbf{MAE} & \textbf{conf. MSE} & \textbf{conf. MAE} \\
        \midrule
        \multirow{4}{*}{\rotatebox{90}{\textbf{ETTh1}}} 
        & 96  & 0.367 $\pm$ 0.001 & 0.311 $\pm$ 0.000 & 99.73 & 99.99 \\
        & 192 & 0.428 $\pm$ 0.000 & 0.337 $\pm$ 0.000 & 99.99 & 99.99 \\
        & 336 & 0.475 $\pm$ 0.002 & 0.355 $\pm$ 0.001 & 99.58 & 99.72 \\
        & 720 & 0.464 $\pm$ 0.001 & 0.373 $\pm$ 0.002 & 99.78 & 99.46 \\
        \midrule
        \multirow{4}{*}{\rotatebox{90}{\textbf{ETTh2}}} 
        & 96  & 0.054 $\pm$ 0.000 & 0.144 $\pm$ 0.000 & 99.99 & 99.99 \\
        & 192 & 0.066 $\pm$ 0.001 & 0.161 $\pm$ 0.000 & 98.48 & 99.99 \\
        & 336 & 0.078 $\pm$ 0.000 & 0.178 $\pm$ 0.001 & 99.99 & 99.44 \\
        & 720 & 0.078 $\pm$ 0.001 & 0.183 $\pm$ 0.001 & 98.72 & 99.45 \\
        \midrule
        \multirow{4}{*}{\rotatebox{90}{\textbf{ETTm1}}} 
        & 96  & 0.293 $\pm$ 0.001 & 0.273 $\pm$ 0.000 & 99.66 & 99.99 \\
        & 192 & 0.343 $\pm$ 0.000 & 0.296 $\pm$ 0.001 & 99.99 & 99.66 \\
        & 336 & 0.383 $\pm$ 0.001 & 0.318 $\pm$ 0.001 & 99.74 & 99.69 \\
        & 720 & 0.465 $\pm$ 0.001 & 0.354 $\pm$ 0.000 & 99.78 & 99.99 \\
        \midrule
        \multirow{4}{*}{\rotatebox{90}{\textbf{ETTm2}}} 
        & 96  & 0.037 $\pm$ 0.001 & 0.116 $\pm$ 0.000 & 97.30 & 99.99 \\
        & 192 & 0.048 $\pm$ 0.000 & 0.130 $\pm$ 0.000 & 99.99 & 99.99 \\
        & 336 & 0.059 $\pm$ 0.001 & 0.146 $\pm$ 0.000 & 98.31 & 99.99 \\
        & 720 & 0.074 $\pm$ 0.001 & 0.166 $\pm$ 0.000 & 98.65 & 99.99 \\  
        \midrule
        \multirow{4}{*}{\rotatebox{90}{\textbf{Electricity}}} 
        & 96  & 0.017 $\pm$ 0.000 & 0.013 $\pm$ 0.001 & 99.99 & 92.31 \\
        & 192 & 0.023 $\pm$ 0.001 & 0.014 $\pm$ 0.001 & 95.65 & 92.86 \\
        & 336 & 0.038 $\pm$ 0.001 & 0.019 $\pm$ 0.000 & 97.37 & 99.99 \\
        & 720 & 0.045 $\pm$ 0.000 & 0.026 $\pm$ 0.001 & 99.99 & 96.15 \\
        \midrule
        \multirow{4}{*}{\rotatebox{90}{\textbf{Traffic}}} 
        & 96  & 0.126 $\pm$ 0.000 & 0.085 $\pm$ 0.000 & 99.99 & 99.99 \\
        & 192 & 0.207 $\pm$ 0.000 & 0.110 $\pm$ 0.001 & 99.99 & 99.09 \\
        & 336 & 0.279 $\pm$ 0.001 & 0.182 $\pm$ 0.001 & 99.64 & 99.45 \\
        & 720 & 0.373 $\pm$ 0.000 & 0.268 $\pm$ 0.001 & 99.99 & 99.63 \\
        \midrule
        \multirow{4}{*}{\rotatebox{90}{\textbf{Weather}}} 
        & 96  & 0.026 $\pm$ 0.000 & 0.045 $\pm$ 0.000 & 99.99 & 99.99 \\
        & 192 & 0.026 $\pm$ 0.000 & 0.048 $\pm$ 0.000 & 99.99 & 99.99 \\
        & 336 & 0.027 $\pm$ 0.001 & 0.049 $\pm$ 0.000 & 96.30 & 99.99 \\
        & 720 & 0.026 $\pm$ 0.001 & 0.051 $\pm$ 0.001 & 96.15 & 98.04 \\
        \midrule
        \multirow{4}{*}{\rotatebox{90}{\textbf{Exchange}}} 
        & 96  & 0.004 $\pm$ 0.000 & 0.038 $\pm$ 0.000 & 99.99 & 99.99 \\
        & 192 & 0.008 $\pm$ 0.000 & 0.056 $\pm$ 0.000 & 99.99 & 99.99 \\
        & 336 & 0.014 $\pm$ 0.000 & 0.078 $\pm$ 0.001 & 99.99 & 98.72 \\
        & 720 & 0.034 $\pm$ 0.001 & 0.126 $\pm$ 0.000 & 97.06 & 99.99 \\
        \bottomrule
    \end{tabular}
    }
    \label{tab:error_bar_same_seed}
\end{table}

\newpage

\begin{table}[h]
    \centering
    \caption{Displaying the exact results after using different seed values for all the experiments}
    \scriptsize
    \setlength{\tabcolsep}{4pt}
    \renewcommand{\arraystretch}{1.2}
    \noindent\adjustbox{max width=0.48\textwidth,left}{%
    \begin{tabular}{ccc|c|cc|cc|cc|cc|cc|cc}
        \toprule
        \multicolumn{3}{c|}{\multirow{2}{*}{\textbf{Seed Value}}} &
        \multirow{2}{*}{\makecell[l]{\textbf{Dataset} \\ \textbf{\makecell[l]{ \\ Metric}}}} &
        \multicolumn{2}{c|}{\textbf{ETTh1}} &
        \multicolumn{2}{c|}{\textbf{ETTm1}} &
        \multicolumn{2}{c|}{\textbf{Traffic}} &
        \multicolumn{2}{c|}{\textbf{Weather}} &
        \multicolumn{2}{c|}{\textbf{Exchange}} &
        \multicolumn{2}{c}{\textbf{ECL}} \\
        
        \cmidrule(lr){5-6}
        \cmidrule(lr){7-8}
        \cmidrule(lr){9-10}
        \cmidrule(lr){11-12}
        \cmidrule(lr){13-14}
        \cmidrule(lr){15-16}
         & & & & 
        \textbf{MSE} & \textbf{MAE} &
        \textbf{MSE} & \textbf{MAE} &
        \textbf{MSE} & \textbf{MAE} &
        \textbf{MSE} & \textbf{MAE} &
        \textbf{MSE} & \textbf{MAE} &
        \textbf{MSE} & \textbf{MAE} \\
        \midrule
                 
\multirow{4}{*}{} & 
\multirow{4}{*}{2023} & 
\multirow{4}{*}{\textbf{}} & 96  & 
    \textbf{0.367} & \textbf{0.311} & 
    \textbf{0.292} & \textbf{0.272} & 
    \textbf{0.126} & \textbf{0.085} & 
    \textbf{0.025} & \textbf{0.045} & 
    \textbf{0.004} & \textbf{0.038} & 
    \textbf{0.017} & \textbf{0.013} \\
& & & 192 & 
    \textbf{0.428} & \textbf{0.337} & 
    \textbf{0.338} & \textbf{0.294} & 
    \textbf{0.207} & \textbf{0.110} & 
    \textbf{0.026} & \textbf{0.047} & 
    \textbf{0.008} & \textbf{0.056} & 
    \textbf{0.023} & \textbf{0.014} \\
& & & 336 & 
    \textbf{0.475} & \textbf{0.355} & 
    \textbf{0.377} & \textbf{0.316} & 
    \textbf{0.279} & \textbf{0.182} & 
    \textbf{0.027} & \textbf{0.049} & 
    \textbf{0.014} & \textbf{0.078} & 
    \textbf{0.038} & \textbf{0.019} \\
& & & 720 & 
    \textbf{0.464} & \textbf{0.373} & 
    \textbf{0.459} & \textbf{0.351} & 
    \textbf{0.373} & \textbf{0.268} & 
    \textbf{0.026} & \textbf{0.051} & 
    \textbf{0.034} & \textbf{0.126} & 
    \textbf{0.045} & \textbf{0.026} \\

        \midrule
        \multirow{4}{*}{} & 
        \multirow{4}{*}{2025} & 
        \multirow{4}{*}{}     
        & 96  & 
            0.367 & 0.312 & 0.292 & 0.271 & 0.126 & 0.084 & 0.025 & 0.045 & 0.004 & 0.038 & 0.018 & 0.013 \\
        & & & 192 & 
            0.429 & 0.337 & 0.338 & 0.295 & 0.209 & 0.112 & 0.026 & 0.047 & 0.008 & 0.056 & 0.022 & 0.014 \\
        & & & 336 & 
            0.475 & 0.356 & 0.377 & 0.315 & 0.280 & 0.181 & 0.027 & 0.050 & 0.014 & 0.078 & 0.037 & 0.019 \\
        & & & 720 & 
            0.463 & 0.373 & 0.459 & 0.351 & 0.372 & 0.267 & 0.026 & 0.052 & 0.035 & 0.126 & 0.044 & 0.026 \\

        \midrule
        \multirow{4}{*}{} & 
        \multirow{4}{*}{42} & 
        \multirow{4}{*}{} 
        
        & 96  & 
            0.367 & 0.311 & 0.292 & 0.270 & 0.127 & 0.084 & 0.025 & 0.046 & 0.004 & 0.038 & 0.017 & 0.013 \\
        & & & 192 & 
            0.429 & 0.338 & 0.338 & 0.295 & 0.207 & 0.110 & 0.026 & 0.048 & 0.008 & 0.056 & 0.023 & 0.014 \\
        & & & 336 & 
            0.477 & 0.356 & 0.377 & 0.316 & 0.280 & 0.182 & 0.027 & 0.049 & 0.015 & 0.077 & 0.038 & 0.019 \\
        & & & 720 & 
            0.465 & 0.373 & 0.460 & 0.351 & 0.373 & 0.269 & 0.027 & 0.051 & 0.034 & 0.126 & 0.045 & 0.027 \\

        \bottomrule
    \end{tabular}
    }
    \label{tab:error_bar_different_seed_actual_result}
\end{table}

\begin{table}[h]
  \centering
  \caption{Repeating all the experiments with different seeds and reporting MSE, MAE and confidence level}
  \scriptsize
  \setlength{\tabcolsep}{4pt}
  \renewcommand{\arraystretch}{1.2}
  \adjustbox{max width=\columnwidth}{%
  \begin{tabular}{c|c|cc|cc}
    \toprule
    \multirow{2}{*}{\rotatebox{90}{Dataset}} 
      & \multirow{2}{*}{Horizon} 
      & \multicolumn{2}{c|}{Metrics} 
      & \multicolumn{2}{c}{Confidence (\%)} \\
      &               & MSE          & MAE          & MSE      & MAE      \\
    \midrule
    \multirow{4}{*}{\rotatebox{90}{ETTh1}}
      & 96   & 0.367 $\pm$ 0.000 & 0.311 $\pm$ 0.001 & 100.0 &  99.8 \\
      & 192  & 0.428 $\pm$ 0.001 & 0.337 $\pm$ 0.001 &  99.8 &  99.9 \\
      & 336  & 0.475 $\pm$ 0.000 & 0.355 $\pm$ 0.001 & 100.0 &  99.7 \\
      & 720  & 0.464 $\pm$ 0.001 & 0.373 $\pm$ 0.000 &  99.8 & 100.0 \\
    \midrule
    \multirow{4}{*}{\rotatebox{90}{ETTm1}}
      & 96   & 0.292 $\pm$ 0.000 & 0.272 $\pm$ 0.002 & 100.0 &  99.4 \\
      & 192  & 0.338 $\pm$ 0.000 & 0.294 $\pm$ 0.001 & 100.0 &  99.7 \\
      & 336  & 0.377 $\pm$ 0.000 & 0.316 $\pm$ 0.001 & 100.0 &  99.8 \\
      & 720  & 0.459 $\pm$ 0.001 & 0.351 $\pm$ 0.000 &  99.9 & 100.0 \\
    \midrule
    \multirow{4}{*}{\rotatebox{90}{Traffic}}
      & 96   & 0.126 $\pm$ 0.001 & 0.085 $\pm$ 0.001 &  99.6 &  98.8 \\
      & 192  & 0.207 $\pm$ 0.001 & 0.110 $\pm$ 0.001 &  99.5 &  99.1 \\
      & 336  & 0.279 $\pm$ 0.001 & 0.182 $\pm$ 0.001 &  99.6 &  99.7 \\
      & 720  & 0.373 $\pm$ 0.001 & 0.268 $\pm$ 0.001 &  99.9 &  99.6 \\
    \midrule
    \multirow{4}{*}{\rotatebox{90}{Weather}}
      & 96   & 0.025 $\pm$ 0.000 & 0.045 $\pm$ 0.001 & 100.0 &  98.9 \\
      & 192  & 0.026 $\pm$ 0.000 & 0.047 $\pm$ 0.001 & 100.0 &  98.9 \\
      & 336  & 0.027 $\pm$ 0.000 & 0.049 $\pm$ 0.001 & 100.0 &  99.0 \\
      & 720  & 0.026 $\pm$ 0.001 & 0.051 $\pm$ 0.001 &  98.1 &  99.0 \\
    \midrule
    \multirow{4}{*}{\rotatebox{90}{Exchange}}
      & 96   & 0.004 $\pm$ 0.000 & 0.038 $\pm$ 0.000 & 100.0 & 100.0 \\
      & 192  & 0.008 $\pm$ 0.000 & 0.056 $\pm$ 0.000 & 100.0 & 100.0 \\
      & 336  & 0.014 $\pm$ 0.000 & 0.078 $\pm$ 0.001 &  96.4 &  99.4 \\
      & 720  & 0.034 $\pm$ 0.001 & 0.126 $\pm$ 0.000 &  98.5 & 100.0 \\
    \midrule
    \multirow{4}{*}{\rotatebox{90}{ECL}}
      & 96   & 0.017 $\pm$ 0.000 & 0.013 $\pm$ 0.000 &  97.1 & 100.0 \\
      & 192  & 0.023 $\pm$ 0.001 & 0.014 $\pm$ 0.000 &  97.8 & 100.0 \\
      & 336  & 0.038 $\pm$ 0.001 & 0.019 $\pm$ 0.000 &  97.4 & 100.0 \\
      & 720  & 0.045 $\pm$ 0.001 & 0.026 $\pm$ 0.001 &  98.9 &  98.1 \\
    \bottomrule
  \end{tabular}
  }
  \label{tab:error_bar_different_seed_stats_confidence}
\end{table}

\begin{table*}[h]
  \centering
  \caption{Additional results of the long-term forecasting task, evaluating various models across prediction lengths $S \in \{96,192,336,720\}$. Input length is fixed at 96.}
  \scriptsize
  \setlength{\tabcolsep}{2pt}  
  \renewcommand{\arraystretch}{1.1}
  \resizebox{\textwidth}{!}{%
    \begin{tabular}{@{}c|c*{9}{|cc}@{}}
      \toprule
      \multirow{2}{*}{\rotatebox{90}{\textbf{Dataset}}}
        & \multirow{2}{*}{\makecell[l]{\textbf{Models}\\\textbf{Metric}}}
        & \multicolumn{2}{c}{\textbf{Ours}}
        & \multicolumn{2}{c}{\textbf{\makecell{RLinear\\\cite{li2023revisiting}}}}
        & \multicolumn{2}{c}{\textbf{\makecell{PatchTST\\\cite{nie2022time}}}}
        & \multicolumn{2}{c}{\textbf{\makecell{Crossformer\\\cite{zhang2023crossformer}}}}
        & \multicolumn{2}{c}{\textbf{\makecell{TiDE\\\cite{das2023long}}}}
        & \multicolumn{2}{c}{\textbf{\makecell{TimesNet\\\cite{wu2022timesnet}}}}
        & \multicolumn{2}{c}{\textbf{\makecell{DLinear\\\cite{zeng2023transformers}}}}
        & \multicolumn{2}{c}{\textbf{\makecell{SCINet\\\cite{liu2022scinet}}}}
        & \multicolumn{2}{c}{\textbf{\makecell{FEDformer\\\cite{zhou2022fedformer}}}} \\
      \cmidrule(lr){3-4}   
      \cmidrule(lr){5-6}   
      \cmidrule(lr){7-8}   
      \cmidrule(lr){9-10}  
      \cmidrule(lr){11-12} 
      \cmidrule(lr){13-14} 
      \cmidrule(lr){15-16} 
      \cmidrule(lr){17-18} 
      \cmidrule(lr){19-20} 
      & & MSE & MAE & MSE & MAE & MSE & MAE & MSE & MAE & MSE & MAE & MSE & MAE & MSE & MAE & MSE & MAE & MSE & MAE \\
      \midrule

          \multirow{5}{*}{\textbf{\rotatebox{90}{ETTh1}}} 
        & 96  & \textbf{\textcolor{red}{0.367}} & \textbf{\textcolor{red}{0.311}}  & 0.386 & 0.395 & 0.414 & 0.419 & 0.423 & 0.448 & 0.479 & 0.464 & 0.384 & 0.402 & 0.386 & 0.400 & 0.654 & 0.599 & 0.376 & 0.419 \\

        & 192 & 0.428 & \textbf{\textcolor{red}{0.337}} & 0.437 & 0.424 & 0.460 & 0.445 & 0.471 & 0.474 & 0.525 & 0.492 & 0.436 & 0.429 & 0.437 & 0.432 & 0.719 & 0.631 & \textbf{\textcolor{red}{0.420}} & 0.448 \\

        & 336 & 0.475 & \textbf{\textcolor{red}{0.355}} & 0.479 & 0.446 & 0.501 & 0.466 & 0.570 & 0.546 & 0.565 & 0.515 & 0.491 & 0.469 & 0.481 & \textbf{\textcolor{red}{0.459}} & 0.778 & 0.659 & \textbf{\textcolor{red}{0.459}} & 0.465 \\

        & 720 & \textbf{\textcolor{red}{0.464}} & \textbf{\textcolor{red}{0.373}} & 0.481 & 0.470 & 0.500 & 0.488 & 0.653 & 0.621 & 0.594 & 0.558 & 0.521 & 0.500 & 0.519 & 0.516 & 0.836 & 0.699 & 0.506 & 0.507 \\

        \midrule

        \multirow{5}{*}{\textbf{\rotatebox{90}{ETTh2}}} 
        & 96  & \textbf{\textcolor{red}{0.054}} & \textbf{\textcolor{red}{0.144}} & 0.288 & 0.338 & 0.302 & 0.348 & 0.745 & 0.584 & 0.400 & 0.440 & 0.340 & 0.374 & 0.333 & 0.387 & 0.707 & 0.621 & 0.358 & 0.397 \\

        & 192 & \textbf{\textcolor{red}{0.066}} & \textbf{\textcolor{red}{0.161}}  & 0.374 & 0.390 & 0.388 & 0.400 & 0.877 & 0.656 & 0.528 & 0.509 & 0.402 & 0.414 & 0.477 & 0.476 & 0.860 & 0.689 & 0.429 & 0.439 \\

        & 336 & \textbf{\textcolor{red}{0.078}} & \textbf{\textcolor{red}{0.178}} & 0.415 & 0.426 & 0.426 & 0.433 & 1.043 & 0.731 & 0.643 & 0.571 & 0.452 & 0.452 & 0.594 & 0.541 & 1.000 & 0.744 & 0.496 & 0.487 \\

        & 720 & \textbf{\textcolor{red}{0.078}} & \textbf{\textcolor{red}{0.183}} & 0.420 & 0.440 & 0.431 & 0.446 & 1.104 & 0.763 & 0.874 & 0.679 & 0.462 & 0.468 & 0.831 & 0.657 & 1.249 & 0.838 & 0.463 & 0.474 \\
        
        \midrule

        \multirow{5}{*}{\textbf{\rotatebox{90}{ETTm1}}} 
        & 96  & \textbf{\textcolor{red}{0.293}} & \textbf{\textcolor{red}{0.273}} & 0.355 & 0.376 & 0.329 & 0.367 & 0.404 & 0.426 & 0.364 & 0.387 & 0.338 & 0.375 & 0.345 & 0.372 & 0.418 & 0.438 & 0.379 & 0.419 \\

        & 192 & \textbf{\textcolor{red}{0.343}} & \textbf{\textcolor{red}{0.296}} & 0.391 & 0.392 & 0.367 & 0.385 & 0.450 & 0.451 & 0.398 & 0.404 & 0.374 & 0.387 & 0.380 & 0.389 & 0.439 & 0.450 & 0.426 & 0.441 \\

        & 336 & \textbf{\textcolor{red}{0.383}} & \textbf{\textcolor{red}{0.318}} & 0.424 & 0.415 & 0.399 & 0.410 & 0.532 & 0.515 & 0.428 & 0.425 & 0.410 & 0.411 & 0.413 & 0.413 & 0.490 & 0.485 & 0.445 & 0.459 \\

        & 720 & 0.465 & \textbf{\textcolor{red}{0.354}} & 0.487 & 0.450 & \textbf{\textcolor{red}{0.454}} & 0.439 & 0.666 & 0.589 & 0.487 & 0.461 & 0.478 & 0.450 & 0.474 & 0.453 & 0.595 & 0.550 & 0.543 & 0.490 \\

        \midrule

        \multirow{5}{*}{\textbf{\rotatebox{90}{ETTm2}}} 
        & 96  & \textbf{\textcolor{red}{0.037}} & \textbf{\textcolor{red}{0.116}} & 0.182 & 0.265 & 0.175 & 0.259 & 0.287 & 0.366 & 0.207 & 0.305 & 0.187 & 0.267 & 0.193 & 0.292 & 0.286 & 0.377 & 0.203 & 0.287 \\

        & 192 & \textbf{\textcolor{red}{0.048}} & \textbf{\textcolor{red}{0.130}} & 0.246 & 0.304 & 0.241 & 0.302 & 0.414 & 0.492 & 0.290 & 0.364 & 0.249 & 0.309 & 0.284 & 0.362 & 0.399 & 0.445 & 0.269 & 0.328  \\

        & 336 & \textbf{\textcolor{red}{0.059}} & \textbf{\textcolor{red}{0.146}} & 0.307 & 0.342 & 0.305 & 0.343 & 0.597 & 0.542 & 0.377 & 0.422 & 0.321 & 0.351 & 0.369 & 0.427 & 0.637 & 0.591 & 0.325 & 0.366  \\

        & 720 & \textbf{\textcolor{red}{0.074}} & \textbf{\textcolor{red}{0.166}} & 0.407 & 0.398 & 0.402 & 0.400 & 1.730 & 1.042 & 0.558 & 0.524 & 0.408 & 0.403 & 0.554 & 0.522 & 0.960 & 0.735 & 0.421 & 0.415  \\

        \midrule

        \multirow{5}{*}{\textbf{\rotatebox{90}{Electricity}}} 
        & 96  & \textbf{\textcolor{red}{0.017}} & \textbf{\textcolor{red}{0.013}} & 0.201 & 0.281 & 0.181 & 0.270 & 0.219 & 0.314 & 0.237 & 0.329 & 0.168 & 0.272 & 0.197 & 0.282 & 0.247 & 0.345 & 0.193 & 0.308  \\

        & 192 & \textbf{\textcolor{red}{0.023}} & \textbf{\textcolor{red}{0.014}} & 0.201 & 0.283 & 0.188 & 0.274 & 0.231 & 0.322 & 0.236 & 0.330 & 0.184 & 0.289 & 0.196 & 0.285 & 0.257 & 0.355 & 0.201 & 0.315  \\

        & 336 & \textbf{\textcolor{red}{0.038}} & \textbf{\textcolor{red}{0.019}} & 0.215 & 0.298 & 0.204 & 0.293 & 0.246 & 0.337 & 0.249 & 0.344 & 0.198 & 0.300 & 0.209 & 0.301 & 0.269 & 0.369 & 0.214 & 0.329  \\

        & 720 & \textbf{\textcolor{red}{0.045}} & \textbf{\textcolor{red}{0.026}} & 0.257 & 0.331  & 0.246 & 0.324 & 0.280 & 0.363 & 0.284 & 0.373 & 0.220 & 0.320 & 0.245 & 0.333 & 0.299 & 0.390 & 0.246 & 0.355 \\

        \midrule

        \multirow{5}{*}{\textbf{\rotatebox{90}{Traffic}}} 
        & 96  & \textbf{\textcolor{red}{0.126}} & \textbf{\textcolor{red}{0.085}} & 0.649 & 0.389 & 0.462 & 0.295 & 0.522 & 0.290 & 0.805 & 0.493 & 0.593 & 0.321 & 0.650 & 0.396 & 0.788 & 0.499 & 0.587 & 0.366  \\

        & 192 & \textbf{\textcolor{red}{0.207}} & \textbf{\textcolor{red}{0.110}} & 0.601 & 0.366 & 0.466 & 0.296 & 0.530 & 0.293 & 0.756 & 0.474 & 0.617 & 0.336 & 0.598 & 0.370 & 0.789 & 0.505 & 0.604 & 0.373  \\

        & 336 & \textbf{\textcolor{red}{0.279}} & \textbf{\textcolor{red}{0.182}} & 0.609 & 0.369 & 0.482 & 0.304 & 0.558 & 0.305 & 0.762 & 0.477 & 0.629 & 0.336 & 0.605 & 0.373 & 0.797 & 0.508 & 0.621 & 0.383  \\

        & 720 & \textbf{\textcolor{red}{0.373}} & \textbf{\textcolor{red}{0.268}} & 0.647 & 0.387 & 0.514 & 0.322 & 0.589 & 0.328 & 0.719 & 0.449 & 0.640 & 0.350 & 0.645 & 0.394 & 0.841 & 0.523 & 0.626 & 0.382  \\

        \midrule

        \multirow{5}{*}{\textbf{\rotatebox{90}{Weather}}} 
        & 96  & \textbf{\textcolor{red}{0.026}} & \textbf{\textcolor{red}{0.045}} & 0.192 & 0.232 & 0.177 & 0.218 & 0.158 & 0.230 & 0.202 & 0.261 & 0.172 & 0.220 & 0.196 & 0.255 & 0.221 & 0.306 & 0.217 & 0.296  \\

        & 192 & \textbf{\textcolor{red}{0.026}} & \textbf{\textcolor{red}{0.048}} & 0.240 & 0.271 & 0.225 & 0.259 & 0.206 & 0.277 & 0.242 & 0.298 & 0.219 & 0.261 & 0.237 & 0.296 & 0.261 & 0.340 & 0.276 & 0.336  \\

        & 336 & \textbf{\textcolor{red}{0.027}} & \textbf{\textcolor{red}{0.049}} & 0.292 & 0.307 & 0.278 & 0.297 & 0.272 & 0.335 & 0.287 & 0.335 & 0.280 & 0.306 & 0.283 & 0.335 & 0.309 & 0.378 & 0.339 & 0.380  \\

        & 720 & \textbf{\textcolor{red}{0.026}} & \textbf{\textcolor{red}{0.051}} & 0.364 & 0.353 & 0.354 & 0.348 & 0.398 & 0.418 & 0.351 & 0.386 & 0.365 & 0.359 & 0.345 & 0.381 & 0.377 & 0.427 & 0.403 & 0.428  \\

        \midrule
        
        \multirow{5}{*}{\textbf{\rotatebox{90}{Exchange}}} 
        & 96  & \textbf{\textcolor{red}{0.004}} & \textbf{\textcolor{red}{0.038}} & 0.093 & 0.217 & 0.088 & 0.205 & 0.256 & 0.367 & 0.094 & 0.218 & 0.107 & 0.234 & 0.088 & 0.218 & 0.267 & 0.396 & 0.148 & 0.278  \\

        & 192 & \textbf{\textcolor{red}{0.008}} & \textbf{\textcolor{red}{0.056}} & 0.184 & 0.307 & 0.176 & 0.299 & 0.470 & 0.509 & 0.184 & 0.307 & 0.226 & 0.344 & 0.176 & 0.315 & 0.351 & 0.459 & 0.271 & 0.315  \\

        & 336 & \textbf{\textcolor{red}{0.014}} & \textbf{\textcolor{red}{0.078}} & 0.351 & 0.432 & 0.301 & 0.397 & 1.268 & 0.883 & 0.349 & 0.431 & 0.367 & 0.448 & 0.313 & 0.427 & 1.324 & 0.853 & 0.460 & 0.427  \\

        & 720 & \textbf{\textcolor{red}{0.034}} & \textbf{\textcolor{red}{0.126}} & 0.886 & 0.714 & 0.901 & 0.714 & 1.767 & 1.068 & 0.852 & 0.698 & 0.964 & 0.746 & 0.839 & 0.695 & 1.058 & 0.797 & 1.195 & 0.695  \\

      \bottomrule
    \end{tabular}
  }
  \label{tab:main_table_two}
\end{table*}










\section{Showcases}

\begin{figure*}[t!]
  \centering

  \begin{minipage}{\textwidth}
    \centering
    \includegraphics[width=\textwidth]{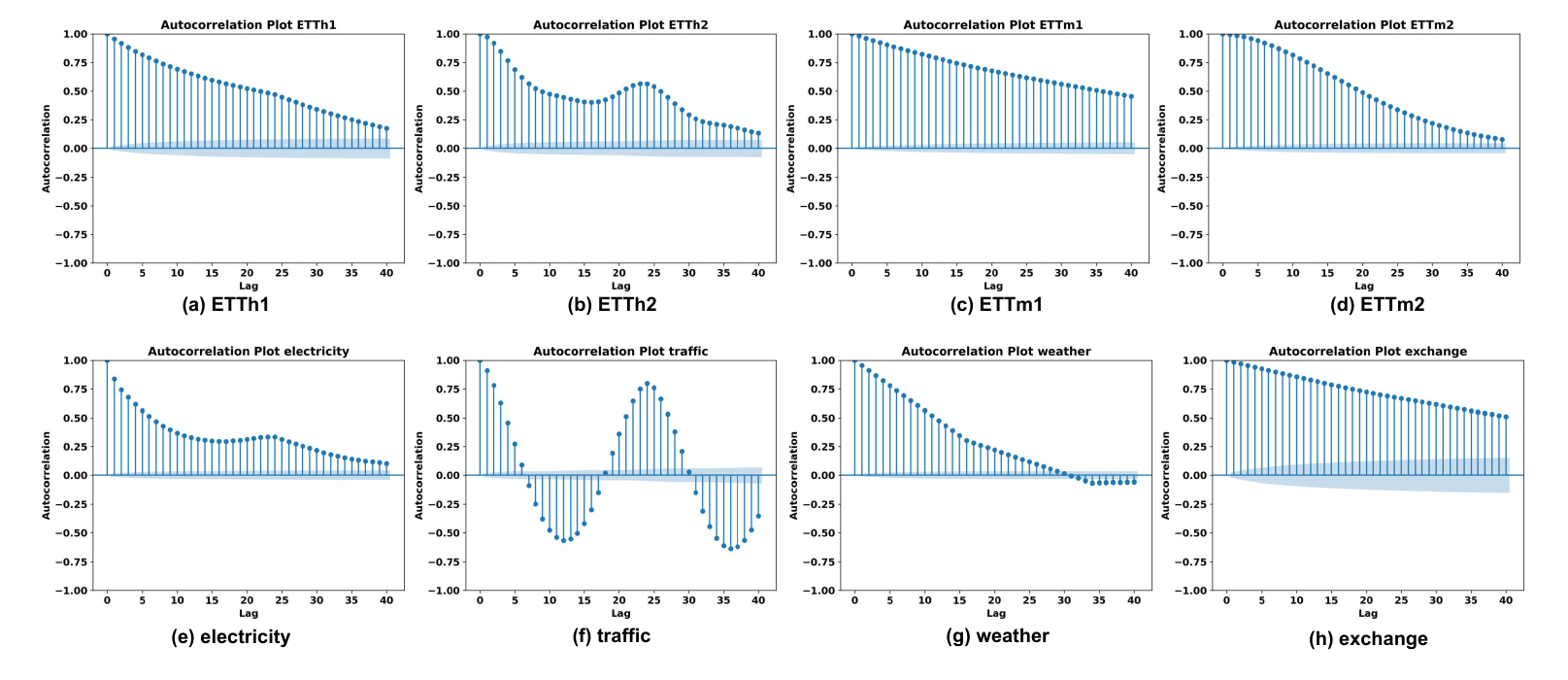}
    \captionof{figure}{Autocorrelation graph for the residual series.}
    \label{fig:autocorrelation}
  \end{minipage}

  \vspace{0.5em}

  \begin{minipage}{\textwidth}
    \centering
    \includegraphics[width=\textwidth]{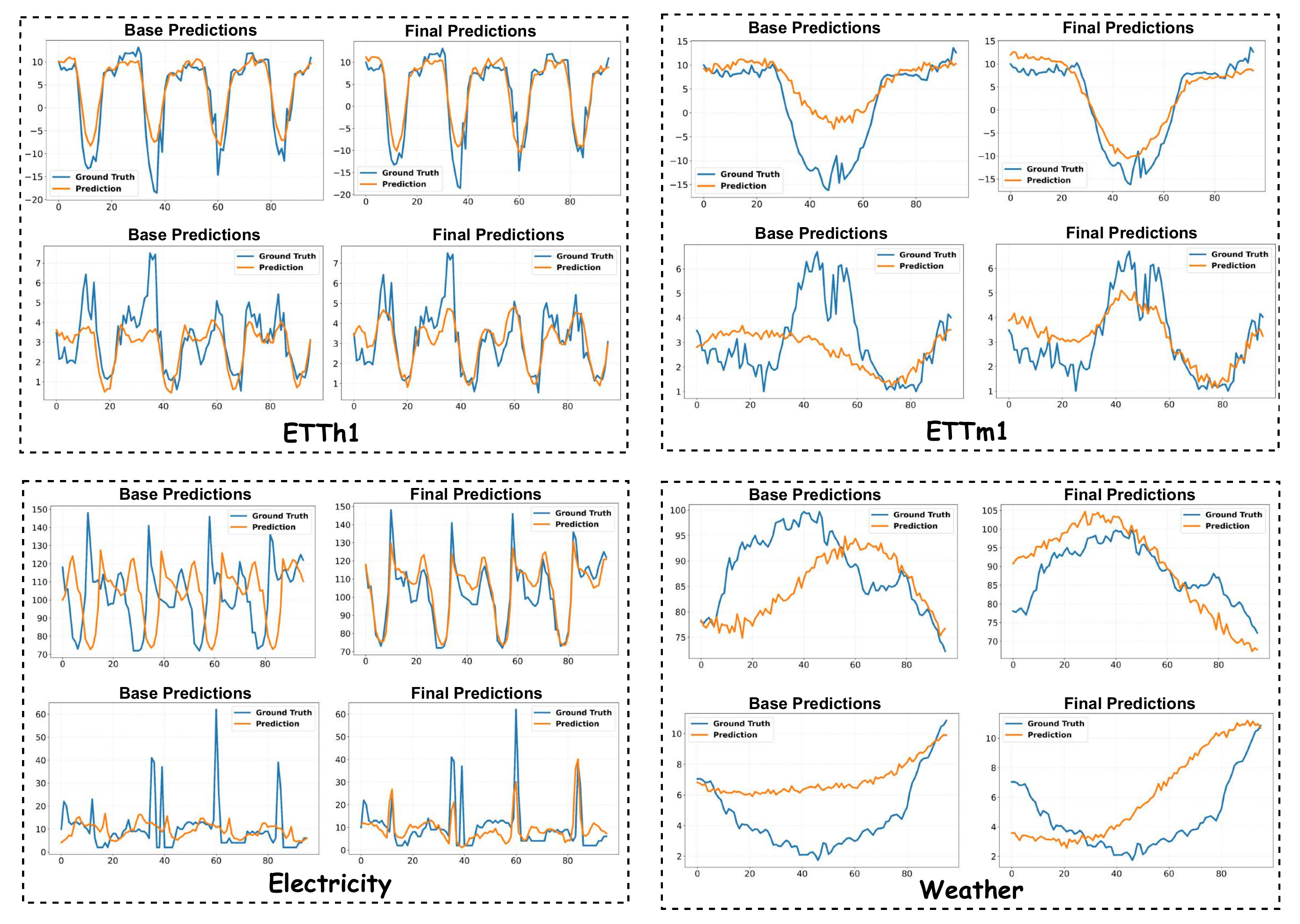}
    \captionof{figure}{Effect of our two-stage pipeline: base predictions generated by $f_{\theta}$ and final predictions obtained by residual calibration using the meta-model $f_{\phi}$.}
    \label{fig:showcase}
  \end{minipage}

\end{figure*}

In this section, we demonstrate the impact of residual correction on the base model’s forecasts. As shown in Figure \ref{fig:showcase}, adjusting the residual predictions brings the base forecasts markedly closer to the ground truth. This result confirms that our proposed pipeline effectively mitigates the structured residual bias of the base model and thereby validates our hypothesis.




\end{document}